\documentclass[journal]{IEEEtran}  
\ifCLASSINFOpdf
\else
\fi
\usepackage{amsmath,graphicx,cite}
\usepackage{subfig}
\usepackage{widetext}
\usepackage{algorithm}
\usepackage{algcompatible}
\usepackage[end]{algpseudocode}
\usepackage{rotating}

\makeatletter
\def\BState{\State\hskip-\ALG@thistlm}
\makeatother

\newcommand{\myindent}[1]{
\newline\makebox[#1cm]{}
}

\begin{document}
%
\title{Development of a N-type GM-PHD Filter for Multiple Target, Multiple Type Visual Tracking}
%
%
%

\author{Nathanael~L.~Baisa ,~\IEEEmembership{Member,~IEEE,}
        and~Andrew~Wallace,~\IEEEmembership{Fellow,~IET}
        \\
\thanks{N. L. Baisa and A. Wallace are with the Department of Electrical, Electronic and Computer Engineering, Heriot Watt University, Edinburgh EH14 4AS, United Kingdom. E-mail: \{nb30, a.m.wallace\}@hw.ac.uk.}
}

\maketitle

\begin{abstract}

We propose a new framework that extends the standard Probability Hypothesis Density (PHD) filter for multiple targets having $N\geq2$ different types based on Random Finite Set theory, taking into account not only background clutter, but also confusions among detections of different target types, which are in general different in character from background clutter. Under Gaussianity and linearity assumptions, our framework extends the existing Gaussian mixture (GM) implementation of the standard PHD filter to create a N-type GM-PHD filter. The methodology is applied to real video sequences by integrating object detectors' information into this filter for two scenarios. For both cases, Munkres's variant of the Hungarian assignment algorithm is used to associate tracked target identities between frames. This approach is evaluated and compared to both raw detection and independent GM-PHD filters using the Optimal Sub-pattern Assignment metric and discrimination rate. This shows the improved performance of our strategy on real video sequences.


\end{abstract}

\begin{IEEEkeywords}
Visual tracking, Random finite sets, FISST, Multiple target filtering, PHD filter, N-type GM-PHD filter, Gaussian mixture, OSPA metric
\end{IEEEkeywords}

%
\IEEEpeerreviewmaketitle

\section{Introduction}

Multi-target tracking, one type or multi-type, is an active research field in computer vision with a wide variety of applications such as intelligent surveillance, human-computer (robot) interaction, augmented reality, visual servoing, robot navigation and autonomous driving. It essentially associates the detections corresponding to the same object over time i.e. it assigns consistent labels to the tracked targets in each video frame to generate a trajectory for each target. It also estimates the number of targets (cardinality) in the scene. These can be performed using online~\cite{SanPoiCav16}\cite{SonJeo16} or offline~\cite{LeaCanSch16}\cite{MilRotSch14}\cite{PirRamFow11} approaches. Online methods estimate the target state at each time instant and depend on predictive models in the case of missed detections to carry on tracking. In contrast, offline (batch) methods use both past and future observations to overcome missed detections. Although offline trackers can generally outperform the online trackers, they are limited for real-time applications.

Using a target detection process applied to the source data, a multi-target tracker receives a random number of measurements caused by such detection uncertainty. Thus, the multi-target tracker has to deal with measurement origin uncertainty, false alarms, missed detections, and the births and deaths of targets in addition to process and measurement noises. Comprehensive surveys of multi-target tracking methods can be found in~\cite{LuoZhaKim14} and \cite{VoMalBarCorOsbMahVo15}. In particular, three data association algorithms, Global Nearest Neighbor (GNN)~\cite{CaiFreLit06}, Joint Probabilistic Data Association Filter (JPDAF)~\cite{RasHag01}, and Multiple Hypothesis Tracking (MHT)~\cite{ChaReh99} have been widely used for more than three decades.
All of these methods have significant complexity, although the performance and computational cost of the MHT is much higher than that of the GNN or JPDAF. In part due to the complexity of these approaches, random finite set (RFS) multi-target tracking
algorithms~\cite{Mah14} have received a great deal of attention. This paradigm includes all sources of uncertainty in a unified framework.
The probability hypothesis density (PHD) filter~\cite{Mah03} is the most widely adopted RFS-based filter
for visual tracking of targets.

In many computer vision applications, for example for situational awareness, driver assistance and vehicle autonomy, there is also a necessity to distinguish between different target types, e.g. between vehicles and more vulnerable road users such as pedestrians and bicycles to select the best sensor focus and course of action~\cite{Matzka2012}.
In sports analysis we often want to track and discriminate sub-groups of the same target type such as players in opposing teams~\cite{LiuCarr14}.
In this and many other examples, confusion between target types is common; a standard histogram-based detection strategy~\cite{DolAppPerBel14} in an urban environment may provide confused detections between pedestrians and cyclists, and even small cars.


In this work, we mainly focus on extending the standard PHD filter based on RFS theory for multiple targets having $N\geq2$ different types (classes), taking into account not only background clutter, but also confusions among detections of different target types at the measurement stage.
We also extend the existing Gaussian mixture (GM) implementation of the standard PHD filter to create a N-type GM-PHD filter using assumptions of Gaussianity and linearity.
Furthermore, we apply the methodology to real video sequences by integrating the object detectors' information into this filter for two application areas.
In each case, we employ Munkres's variant of the Hungarian assignment algorithm to associate tracked target identities between frames.

The main contributions of this paper are as follows.
\begin{itemize}
  \item We model RFS filtering of $N$ different types of multiple targets with separate but confused detections where $N\geq2$.
  \item The Gaussian mixture implementation of the standard PHD filter~\cite{VoMa06} is extended for the proposed N-type PHD filter.
  \item We train and apply object detectors to video sequences of a soccer game, using players from each team and the referee as three distinct target types, and urban scenes, using pedestrians and vehicles as two distinct target types. Compared to ground truth we extract the detection, confusion and background clutter probabilities and integrate these into the N-type GM-PHD filter.
  \item We apply Munkres's variant of the Hungarian assignment algorithm to the typed results from the N-type GM-PHD filter to determine individual targets of each type between consecutive frames.
	\item We compare our approach to both repeated detection and the standard $N$ independent GM-PHD filters to show that our approach yields improved performance in both target identification and location.
\end{itemize}

We have presented preliminary ideas in~\cite{BaiWal17} (simulation under varying probabilities of confusion) and \cite{NatAnd17} (on video application).
In this paper we further develop the theoretical approach, extending from a tri-PHD filter to a N-type PHD filter.
We conduct experiments on tracking vehicles and pedestrians, as two different target types, and
on tracking two football teams and a referee, as three different target types within video sequences.
We also evaluate our results using the Multi-object Tracking (MOT) benchmark.

The remainder of this paper is organized as follows.
In section~\ref{Sec:ReladWork}, we discuss related work.
Multiple type, multiple target recursive Bayes filtering with RFS is described in section~\ref{Sec:FilteringWithRFS}.
A probability generating functional for deriving the N-type PHD filter and the N-type PHD filtering strategy are given in sections~\ref{Sec:PGFL} and~\ref{Sec:DualPHDfilter} respectively.
In section~\ref{Subsec:GM-DualPHD}, a Gaussian mixture implementation of the N-type PHD filter is described in detail. The experimental results are analyzed and compared in section~\ref{Sec:ExperimentalResults}.
The main conclusions and suggestions for future work are summarized in section~\ref{Sec:Conclusion}.

\section{Related work} \label{Sec:ReladWork}

Traditionally, multi-target trackers including GNN~\cite{CaiFreLit06}, JPDAF~\cite{RasHag01}, and MHT~\cite{ChaReh99},
are based on the concept of finding associations between targets and measurements.
However, these approaches have faced challenges not only in the uncertainty caused by data association but also in algorithmic complexity that increases exponentially with the number of targets and measurements. For instance, the total number of possible hypotheses in MHT increases exponentially with time and heuristic pruning/merging of hypotheses is performed to reduce computational cost.

To address the problems of increasing complexity, a unified framework that directly extends single to multiple target tracking by representing multi-target states and observations as RFS was developed by Mahler~\cite{Mah03}.
This estimates the states and cardinality of an unknown and time varying number of targets in the scene, and allows for target birth, death, handles clutter (false alarms) and missing detections.
Mahler propagated the first-order moment of the multi-target posterior, called the Probability Hypothesis Density (PHD), rather than the full multi-target posterior.

There are two popular implementation schemes for the PHD filter, the Gaussian mixture (GM-PHD)~\cite{VoMa06} and the Sequential Monte Carlo (SMC) or particle-PHD filter~\cite{VoSinDou05}.
The GM-PHD filter is preferred for linear (and by extension mildly non-linear) dynamic and observation models and assumes a Gaussian stochastic process~\cite{VoMa06}.
However, for highly non-linear dynamic and observation models and non-Gaussian stochastic process, the SMC-PHD filter is the better implementation scheme~\cite{VoSinDou05}.
For example, the GM-PHD filter is used in~\cite{ZhoLi14} for tracking pedestrians in video sequences but there is only one type of target and the motion model is fixed, and in~\cite{BaiBhoWal17}\cite{BaiBhoWal18} for selective tracking in sparse and dense environments.
As an extension, a GM-PHD Filter was also developed in~\cite{PasVoTuaMa09} for maneuvering targets but this employed a Jump Markov System (JMS) that switched between several motion models.
In contrast, a particle-PHD filter was applied in~\cite{MagTajCav08} to allow for more complex motion models, and to cope with variation of scale, which has significant effects not just on object motion but also on the detection process. It was also used in~\cite{SanPoiCav16}, treating high-confidence (strong) and low-confidence (weak) detections separately for better performance.

Considering extensions to different target types, Yan et al.~\cite{YanFuLonLi12} developed detection, tracking and classification (JDTC) of multiple targets in clutter by jointly estimating the number of targets, their kinematic states, and types of targets (classes) from a sequence of noisy and cluttered observation sets using a SMC-PHD filter.
The dynamics of each target type (class) was modeled as a class-dependent model set and the signal amplitude was included in the multi-target likelihood to enhance the discrimination between targets from different classes and false alarms.
Similarly, a joint target tracking and classification (JTC) algorithm was developed in~\cite{YanFuLi14} using RFS which takes into account extraneous target-originated measurements (of the same type), i.e. multiple measurements that originated from a target which can be modeled as a Poisson RFS using linear and Gaussian assumptions.
In these approaches, the augmented state vector of a target comprises the target kinematic state and class label, i.e. the target type (class) is put into the target state vector.
Simultaneous multi-object tracking and classification was proposed by~\cite{RomAgaNie16} using a graphical probabilistic model and an inference procedure, then solving using a variational approximation, but this
approach suffers from class switching.
Although multiple target types were considered, no account was taken of the effect of confusion between target types at the detection stage, as is the case in our work.

\section{Multiple Target, Multiple Type Recursive Bayes Filtering with RFS} \label{Sec:FilteringWithRFS}

A RFS represents a varying number of non-ordered target states and observations, analogous to a random vector for single target tracking.
More precisely, a RFS is a finite-set-valued random variable i.e. a random variable which is random in both the number of elements and the values of the elements themselves.
Finite Set Statistics (FISST), the study of the statistical properties of RFS, is a systematic treatment of multi-sensor multi-target filtering as a unified Bayesian framework using random set theory~\cite{Mah03}.

When different detectors run on the same scene to detect different target types there is no guarantee that these detectors only detect their own type.
It is possible to run an independent PHD filter for each target type, but this
will not be correct in most cases, as the likelihood of a positive response to a target of the wrong
type will in general be different from, usually higher than, the likelihood of a positive response within the scene background (false alarm).
In this paper, we account for this difference between background clutter and target type confusion.
This is equivalent to a single sensor (e.g. a smart camera) that has $N$ different detection modes, each with its own probability of detection and a measurement density for $N$ different target types.


To derive the N-type PHD filter, we define a RFS representation that extends from a single type, single-target Bayes framework to a multiple type, multiple target Bayes framework.
Let the multi-target state space $\mathcal{F}(\mathcal{X})$ and the multi-target observation space $\mathcal{F}(\mathcal{Z})$ be the respective collections of all the finite subsets of the state space $\mathcal{X}$ and observation space $\mathcal{Z}$, respectively.
If $L_i(k)$ is the number of targets of target type $i$ in the scene at time $k$, then the multiple states for target type $i$, $X_{i,k}$, is the set

\begin{equation}
    X_{i,k} =  \{x_{i,k,1},...x_{i,k,L_i(k)}\} \in \mathcal{F}(\mathcal{X})
\label{eqn:stateSet1}
\end{equation}
\noindent where $i \in \{1, ..., N\}$. Similarly, if $M_j(k)$ is the number of received observations from detector $j$, then the corresponding multiple target measurements is the set

\begin{equation}
    Z_{j,k} =  \{z_{j,k,1},...z_{j,k,M_j(k)}\} \in \mathcal{F}(\mathcal{Z})
\label{eqn:observationSet1}
\end{equation}
\noindent where $j \in \{1, ..., N\}$.
As stated above, some of these observations may be false, i.e. due to clutter (background) or confusion (response due to another target type).

The uncertainty in the state and measurement is introduced by modeling the multi-target state and the multi-target measurement using RFS.
Let $\Xi_{i,k}$ be the RFS associated with the multi-target state of target type $i$, then

\begin{equation}
    \Xi_{i,k} =  S_{i,k}(X_{i,k-1}) \cup \Gamma_{i,k},
\label{eqn:stateRFS1}
\end{equation}
\noindent where $S_{i,k}(X_{i,k-1})$ denotes the RFS of surviving targets of target type $i$, and $\Gamma_{i,k}$ is the RFS of new-born targets of target type $i$. We do not consider spawned targets as these have no meaning in our context, discussed below.

Further, the RFS $\Omega_{ji,k}$ associated with the multi-target measurements of target type $i$ from detector $j$ is
\begin{equation}
    \Omega_{ji,k} =  \Theta_{j,k}(X_{i,k}) \cup C_{si,k} \cup C_{tiJ,k},
\label{eqn:observationRFS1}
\end{equation}
\noindent where $J=\{1, ..., N\}\setminus i$ and $\Theta_{j,k}(X_{i,k})$ is the RFS modeling the measurements generated by the targets $X_{i,k}$, and $C_{s_{i,k}}$ models the RFS associated with the clutter (false alarms) for target type $i$ which comes from the scene background. However, we also include $C_{t_{iJ,k}}$ which is the RFS associated with measurements of all target types $J=\{1, ..., N\}\setminus i$, that is confusions while filtering target type $i$.

Analogous to the single-target case, the dynamics of $\Xi_{i,k}$ are described by the multi-target transition density $y_{i,k|k-1}(X_{i,k}|X_{i,k-1})$, while $\Omega_{ji,k}$ is described by the multi-target likelihood $f_{ji,k}(Z_{j,k}|X_{i,k})$ for target type $i \in \{1, ..., N\}$ from detector $j \in \{1, ..., N\}$. The recursive equations are

\begin{equation}
\begin{array} {lll}
p_{i,k|k-1}(X_{i,k}|Z_{j,1:k-1}) = \\ \int y_{i,k|k-1}(X_{i,k}|X)p_{i,k-1|k-1}(X|Z_{j,1:k-1})\mu(dX)
\end{array}
\label{eqn:predictionRFS}
\end{equation}
\noindent
\begin{equation}
\begin{array} {lll}
    p_{i,k|k}(X_{i,k}|Z_{j,1:k}) =  \frac {f_{ji,k}(Z_{j,k}|X_{i,k})p_{i,k|k-1}(X_{i,k}|Z_{j,1:k-1})}{\int f_{ji,k}(Z_{j,k}|X)p_{i,k|k-1}(X|Z_{j,1:k-1})\mu(dX)}
\end{array}
\label{eqn:updateRFS}
\end{equation}
\noindent where $\mu$ is an appropriate dominating measure on $\mathcal{F}(\mathcal{X})$~\cite{Mah03}. Though a Monte Carlo approximation of this optimal multi-target types Bayes recursion is possible according to multi-target for single type~\cite{VoSinDou05}, the number of particles required is exponentially related to the number of targets and their types in the scene.
To make it computationally tractable, we extend Mahler's method of propagating the first-order moment (PHD) of the multi-target posterior instead of the full multi-target posterior for $N\geq2$ types of multiple targets.
We derive the updated PHDs from Probability Generating Functionals (PGFLs) starting from the standard predicted PHDs of each target type, denoting this as the N-type PHD filter.

\section{Probability Generating Functional (PGFL)} \label{Sec:PGFL}

A probability generating functional is a convenient representation for stochastic modelling with a point process~\cite{Mah03}, a type of random process for which any one realisation consists of a set of isolated points either in time or space. Now, we model joint (probability generating) functionals which take into account the clutter due to the other target types (confusion) in addition to the background clutter for deriving the updated PHDs. Starting from the standard proved predicted PHDs~\cite{Mah03}, we derive novel extensions for the updated PHDs of a N-type PHD filter from PGFLs of each target type, handling confusions among target types.

The joint functional for target type $i$ treating all other target types as clutter is given by

\begin{equation}
    F_i[g,h] = G_{T_i}(hG_{L_{i,i}}(g|.))G_{c_i}(g)\prod_{j=1\setminus i}^N G_{T_j}(G_{L_{j,i}}(g|.)),
\label{eqn:JointFunctional1}
\end{equation}
\noindent where $i \in \{1, ..., N\}$ denotes target type, $g$ is related to the target measurement process and $h$ is related to the target state process.
\begin{equation}
    G_{c_i}(g) = \exp(\lambda_i(c_i[g] - 1)),
\label{eqn:PGFLclutter1}
\end{equation}
\noindent where $G_{c_i}(g)$ is the Poisson PGFL~\cite{Mah03} for false alarms where $\lambda_i$ is the average number of false alarms for target type $i$ and the functional $c_i[g] = \int g(z)c_i(z)dz$ where $c_i(.)$ is the uniform density over the surveillance region;
\begin{equation}
    G_{T_i}(h) = \exp(\mu_i(s_i[h] - 1)),
\label{eqn:PGFLprior}
\end{equation}
\noindent where $G_{T_i}(h)$ is the prior PGFL and $\mu_i$ is the average number of targets, each of which is distributed according to $s_i(x)$ for target type $i$; and

\begin{equation}
    G_{L_{j,i}}(g|x) = 1 - p_{ji,D}(x) + p_{ji,D}(x)\int g(z) f_{ji}(z|x)dz,
\label{eqn:PGFLdetection}
\end{equation}
\noindent where $G_{L_{j,i}}(g|x)$ is the Bernoulli detection process for each target of target type $i$ using detector $j$ with probability of detection for target type $i$ by detector $j$, $p_{ji,D}$, and $f_{ji}(z|x)$ is a likelihood defining the probability that $z$ is generated by the target type $i$ conditioned on state $x$ from detector $j$~\cite{Mah03}.
We expand $s_i[hG_{L_{i,i}}(g|x)]$ and $s_j[G_{L_{j,i}}(g|x)]$ as
\begin{equation}
\begin{array} {lll}
    s_i[hG_{L_{i,i}}(g|x)] =& \displaystyle\int s_i(x) h(x)\Big(1 - p_{ii,D}(x) + \\& p_{ii,D}(x)\int g(z) f_{ii}(z|x)dz\Big)dx,
\end{array}
\label{eqn:PGFLdetectionF1}
\end{equation}
\noindent and
\begin{equation}
\begin{array} {lll}
    s_j[G_{L_{j,i}}(g|x)] =& \displaystyle\int s_j(x)\Big(1 - p_{ji,D}(x) + \\& p_{ji,D}(x)\int g(z) f_{ji}(z|x)dz\Big)dx.
\end{array}
\label{eqn:PGFLdetectionF2}
\end{equation}
\noindent Accordingly, $F_i[g,h]$ is expanded as
\begin{widetext}
\begin{equation}
\begin{array} {lll}
    F_i[g,h] =&  \exp \Bigg( \lambda_i\Big(\int g(z)c_i(z)dz - 1\Big) +  \sum_{j=1\setminus i}^N \mu_j\Big[\displaystyle\int s_j(x)\Big(1 - p_{ji,D}(x) + p_{ji,D}(x)\int g(z) f_{ji}(z|x)dz\Big)dx - 1\Big] + \\&  \mu_i\Big[\displaystyle\int s_i(x) h(x)\Big(1 - p_{ii,D}(x) + p_{ii,D}(x) \int g(z) f_{ii}(z|x)dz\Big)dx - 1\Big] \Bigg) ,
\end{array}
\label{eqn:JointFunctional1Expanded}
\end{equation}
\end{widetext}
\noindent

The updated PGFL $G_i(h|z_1,...z_{M_j})$ for target type $i$ is obtained by finding the $M_j^{th}$ functional derivative of $F_i[g,h]$~\cite{Mah03} and is given by

\begin{equation}
    G_i(h|z_1,...,z_{M_j}) = \frac{\frac{\delta^{M_j}}{\delta_{\varphi_{z_1}}...\delta_{\varphi_{z_{M_j}}}} F_i[g,h]|_{g=0}} {\frac{\delta^{M_j}}{\delta_{\varphi_{z_1}}...\delta_{\varphi_{z_{M_j}}}} F_i[g,1]|_{g=0}},
\label{eqn:updatedPGFL}
\end{equation}
\noindent

The updated PHD for target type $i$ treating all other target types as clutter can be obtained by taking the first-order moment (mean)~\cite{Mah03} of Eq.~(\ref{eqn:updatedPGFL}) and setting $h = 1$,
\begin{widetext}
\begin{equation}
\begin{array} {lll}
\mathcal{D}_i(x|z_1,...,z_{M_j}) &= \frac{\delta}{\delta_{\varphi_x}} G_i(h|z_1,...z_{M_j})|_{h = 1}, \\
                             &= \mu_i s_i(x)(1 - p_{ii,D}(x)) + 
\sum_{m=1}^{M_j} \frac{\mu_i s_i(x)p_{ii,D}(x)f_{ii}(z_m|x)}{\lambda_i c_i(z_m) + \sum_{j=1\setminus i}^N \mu_j \int s_j(x) p_{ji,D}(x) f_{ji}(z_m|x)dx + \mu_i\int s_i(x)p_{ii,D}(x)f_{ii}(z_m|x)dx},
\end{array}
\label{eqn:updatedPHD1}
\end{equation}
\end{widetext}
\noindent This, $\mathcal{D}_i(x|z_1,...,z_{M_j})$ in Eq.~(\ref{eqn:updatedPHD1}), is the updated PHD for target type $i$ treating all other target types as clutter in the case of the N-type PHD filter. The term $\mu_i s_i(x)$ in Eq.~(\ref{eqn:updatedPHD1}) is the predicted PHD for target type $i$.

\section{N-type PHD Filtering Strategy} \label{Sec:DualPHDfilter}

The PHDs, $\mathcal{D}_{\Xi_1}(x)$, $\mathcal{D}_{\Xi_2}(x)$,..., $\mathcal{D}_{\Xi_N}(x)$, are the first-order moments of RFSs, $\Xi_1$, $\Xi_2$,..., $\Xi_N$, and are intensity functions on a single state space $\mathcal{X}$ whose peaks identify the likely positions of the targets. For any region $R\subseteq \mathcal{X}$

\begin{equation}
    E[|(\Xi_1 \cup \Xi_2... \cup\Xi_N) \cap R|] = \sum_{i=1}^N  \int_R \mathcal{D}_{\Xi_i}(x)dx
\label{eqn:PHDcardinality}
\end{equation}

\noindent where$|.|$ is used to denote the cardinality of a set. In practice, Eq.~(\ref{eqn:PHDcardinality}) means that by integrating the PHDs on any region $R$ of the state space,
it is possible to obtain the expected number of targets (cardinality) in $R$.

%
%

At any time step, $k$, new targets may appear (births) and are added to those targets that persist and have moved position from the previous time step.
Consequently, the PHD $\textit{prediction}$ for target type $i$ at time $k$ is
\begin{equation}
\begin{array} {lll}
    \mathcal{D}_{i,k|k-1}(x) = & \int p_{i,S,k|k-1}(\zeta)y_{i,k|k-1}(x|\zeta)\mathcal{D}_{i,k-1|k-1}(\zeta)d\zeta \\& + \gamma_{i,k}(x),
\end{array}
\label{eqn:PHDpredictioni}
\end{equation}
\noindent where $\gamma_{i,k}(.)$ is the intensity function of a new target birth RFS $\Gamma_{i,k}$, $p_{i,S,k|k-1}(\zeta)$ is the probability that a target still exists at time $k$, $y_{i,k|k-1}(.|\zeta)$ is the single target state transition density at time $k$ given the previous state $\zeta$ for target type $i$.

Thus, following Eq.~(\ref{eqn:updatedPHD1}), the final updated PHD for target type $i$ is obtained by setting $\mu_is_i(x) = \mathcal{D}_{i,k|k-1}(x)$

\begin{widetext}
\begin{equation}
\begin{array} {lll}  \mathcal{D}_{i,k|k}(x) &= \bigg[ 1 - p_{ii,D}(x) +
\sum_{z\in Z_{i,k}}\frac{p_{ii,D}(x)f_{ii,k}(z|x)}{c_{s_{i,k}}(z) + c_{t_{i,k}}(z) + \int p_{ii,D}(\xi)f_{ii,k}(z|\xi)\mathcal{D}_{i,k|k-1} (\xi)d\xi} \bigg] \mathcal{D}_{i,k|k-1(x)},
\end{array}
\label{eqn:PHDupdatei}
\end{equation}
\end{widetext}
\noindent The clutter intensity $c_{t_{i,k}}(z)$ due to all types of targets $j \in \{1, ..., N\}$ except target type $i$ in Eq.~(\ref{eqn:PHDupdatei}) is given by

\begin{equation}
\begin{array} {lll}
    c_{t_{i,k}}(z) = \sum_{j \in \{1,...,N\}\setminus i} \int p_{ji,D}(y)\mathcal{D}_{j,k|k-1}(y)f_{ji,k}(z|y)dy,
\end{array}
\label{eqn:Clutteri}
\end{equation}
\noindent This means that when filtering target type $i$, all the other target types are included as confusing detections. Eq.~(\ref{eqn:Clutteri}) converts state space to observation space by integrating the PHD estimator $\mathcal{D}_{j,k|k-1}(y)$ and likelihood $f_{ji,k}(z|y)$ which defines the probability that $z$ is generated by detector $j$ conditioned on state $x$ of the target type $i$ taking into account the confusion probability $ p_{ji,D}(y)$, when target type $i$ is detected by detector $j$.

The clutter intensity due to the background for target type $i$, $c_{s_{i,k}}(z)$, in Eq.~(\ref{eqn:PHDupdatei}) is given by
\begin{equation}
    c_{s_{i,k}}(z) = \lambda_i c_i(z) = \lambda_{c_i} A c_i(z),
\label{eqn:Clutterscenei}
\end{equation}
\noindent where $c_i(.)$ is the uniform density over the surveillance region $A$, and $\lambda_{c_i}$ is the average number of clutter returns per unit volume for target type $i$ i.e. $\lambda_{i} = \lambda_{ci}A$.
While the standard PHD filter has linear complexity with the current number of measurements ($m$) and with the current number of targets ($n$) i.e. computational order of $O(mn)$, the N-type PHD filter has linear complexity with the current number of measurements ($m$), with the current number of targets ($n$) and with the total number of target types ($N$) i.e. computational order of $O(mnN)$.

In general, the clutter intensities due to the background for each target type $i$, $c_{s_{i,k}}(z)$, can be different as they depend on the receiver operating characteristic (ROC) curves of the detection processes.
Moreover, the probabilities of detection $p_{ii,D}(x)$ and $p_{ji,D}(x)$ may all be different although assumed constant across both the time and space continua.

\section{N-type PHD Filter Implementation based on a Gaussian Mixture} \label{Subsec:GM-DualPHD}

The Gaussian mixture implementation of the standard PHD (GM-PHD) filter~\cite{VoMa06} is a closed-form solution of the PHD filter that assumes a linear Gaussian system.
In this section, this is extended for the N-type PHD filter by solving Eq.~(\ref{eqn:Clutteri}).
Assuming each target follows a linear Gaussian model,

\begin{equation}
    y_{i,k|k-1}(x|\zeta) =  \mathcal{N}(x;F_{i,k-1}\zeta, Q_{i,k-1})
\label{eqn:linearState1}
\end{equation}
\noindent
\begin{equation}
    f_{ji,k}(z|x) =  \mathcal{N}(z;H_{ji,k} x, R_{ji,k})
\label{eqn:linearObservation1}
\end{equation}
\noindent where $\mathcal{N}(.;m, P)$ denotes a Gaussian density with mean $m$ and covariance $P$; $F_{i,k-1}$ and $H_{ji,k}$ are the state transition and measurement matrices, respectively. $Q_{i,k-1}$ and $R_{ji,k}$ are the covariance matrices of the process and the measurement noises, respectively, where $i \in \{1, ..., N\}$ and $j \in \{1, ..., N\}$.
A measurement driven birth intensity, similar in principle to ~\cite{RisClaVoVo12}, is introduced at each time step, with a non-informative zero initial target velocity.
This choice is preferred to the options of covering the whole state space (random)~\cite{RisClaBa10} or a-priori birth~\cite{VoMa06} and is discussed further in Section~\ref{Sec:ExperimentalResults}.
The intensity of the spontaneous birth RFS is $\gamma_{i,k}(x)$ for target type $i$

\begin{equation}
\begin{split}
     \gamma_{i,k}(x)  =  \sum_{v = 1}^{V_{\gamma_i,k}} w_{i,\gamma,k}^{(v)}\mathcal{N}(x; m_{i,\gamma,k}^{(v)}, P_{i,\gamma,k}^{(v)})
\label{eqn:PHDbirthassumption2}
\end{split}
\end{equation}
\noindent where $V_{\gamma_i,k}$ is the number of birth Gaussian components for target type $i$ where $i \in \{1,..., N\}$, $m_{i,\gamma,k}^{(v)}$ is the current measurement and zero initial velocity used as mean and $P_{i,\gamma,k}^{(v)}$ is the birth covariance for Gaussian component $v$ of target type $i$.

It is assumed that the posterior intensity for target type $i$ at time $k-1$ is a Gaussian mixture of the form

\begin{equation}
\begin{split}
     \mathcal{D}_{i,k-1}(x)  =  \sum_{v = 1}^{V_{i,k-1}} w_{i,k-1}^{(v)}\mathcal{N}(x; m_{i,k-1}^{(v)}, P_{i,k-1}^{(v)}),
\label{eqn:PHDposterior1k-1}
\end{split}
\end{equation}

\noindent where $i \in \{1, ..., N\}$ and $V_{i,k-1}$ is the number of Gaussian components of $\mathcal{D}_{i,k-1}(x)$.
Under these assumptions, the predicted intensity at time $k$ for target type $i$ is given following Eq.~(\ref{eqn:PHDpredictioni}) by

\begin{equation}
    \mathcal{D}_{i,k|k-1}(x) = \mathcal{D}_{i,S,k|k}(x) + \gamma_{i,k}(x),
\label{eqn:PHDpredictionI1}
\end{equation}
\noindent where

\begin{equation}
\begin{array} {lll}  \mathcal{D}_{i,S,k|k-1}(x) =& p_{i,S,k} \sum_{v = 1}^{V_{i,k-1}} w_{i,k-1}^{(v)}\mathcal{N}(x; \\& m_{i,S,k|k-1}^{(v)},P_{i,S,k|k-1}^{(v)}), \nonumber
\end{array}
\label{eqn:PHDpredictionSurvival1}
\end{equation}
\noindent
\begin{equation}
 m_{i,S,k|k-1}^{(v)} = F_{i,k-1} m_{1,k-1}^{(v)},  \nonumber
\label{eqn:PHDpredictionSurvivalMean1}
\end{equation}
\noindent
\begin{equation}
 P_{i,S,k|k-1}^{(v)} = Q_{i,k-1} + F_{i,k-1} P_{i,k-1}^{(v)} F^T_{1,k-1},  \nonumber
\label{eqn:PHDpredictionSurvivalCov1}
\end{equation}
\noindent where $p_{i,S,k}$ is the survival rate for target type $i$ and $\gamma_{i,k}(x)$ is given by Eq.~(\ref{eqn:PHDbirthassumption2}).

Since $\mathcal{D}_{i,S,k|k-1}(x)$ and $\gamma_{i,k}(x)$ are Gaussian mixtures, $ \mathcal{D}_{i,k|k-1}(x)$ can be expressed as a Gaussian mixture of the form

\begin{equation}
\begin{split}
     \mathcal{D}_{i,k|k-1}(x)  =  \sum_{v = 1}^{V_{i,k|k-1}} w_{i,k|k-1}^{(v)}\mathcal{N}(x; m_{i,k|k-1}^{(v)},P_{i,k|k-1}^{(v)}),
\label{eqn:PHDpredictionki}
\end{split}
\end{equation}
\noindent where $w_{i,k|k-1}^{(v)}$ is the weight accompanying the predicted Gaussian component $v$ for target type $i$ and $V_{i,k|k-1}$ is the number of predicted Gaussian components for target type $i$ where $i \in \{1, ..., N\}$.

Assuming the probabilities of detection to be constant, the posterior intensity for target type $i$ at time $k$ (updated PHD), considering incorrect detection of target types as confusion, is also a Gaussian mixture which corresponds to Eq.~(\ref{eqn:PHDupdatei}) and is given by
\begin{equation}
\begin{split}
     \mathcal{D}_{i,k|k}(x)  =  (1 - p_{ii,D,k})\mathcal{D}_{i,k|k-1}(x) + \sum_{z\in Z_{i,k}} \mathcal{D}_{i,D,k}(x;z),
\label{eqn:PHDupdateki}
\end{split}
\end{equation}
\noindent where
\begin{equation}
\begin{split}
     \mathcal{D}_{i,D,k}(x;z)  =  \sum_{v = 1}^{V_{i,k|k-1}} w_{i,k}^{(v)}(z) \mathcal{N}(x; m_{i,k|k}^{(v)}(z), P_{i,k|k}^{(v)}), \nonumber
\label{eqn:PHDupdateDetki}
\end{split}
\end{equation}
\noindent
\begin{equation}
\begin{split}
     w^{(v)}_{i,k}(z)  =  \frac{p_{ii, D,k} w^{(v)}_{i,k|k-1} q^{(v)}_{i,k}(z)}{c_{s_{i,k}}(z) + c_{t_{i,k}}(z) + p_{ii, D,k} \sum_{l = 1}^{V_{i,k|k-1}} w^{(l)}_{i,k|k-1} q^{(l)}_{i,k}(z)}, \nonumber
\label{eqn:PHDupdatewwightki}
\end{split}
\end{equation}
\noindent
\begin{equation}
\begin{split}
     q^{(v)}_{i,k}(z)  =  \mathcal{N}(z; H_{ii,k} m_{i,k|k-1}^{(v)}, R_{ii,k} + H_{ii,k}P_{i,k|k-1}^{(v)} H^T_{ii,k}), \nonumber
\label{eqn:PHDupdateqki}
\end{split}
\end{equation}
\noindent
\begin{equation}
\begin{split}
     m^{(v)}_{i,k|k}(z)  =  m^{(v)}_{i,k|k-1} + K^{(v)}_{i,k} (z - H_{ii,k} m_{i,k|k-1}^{(v)}), \nonumber
\label{eqn:PHDupdatemki}
\end{split}
\end{equation}
\noindent
\begin{equation}
\begin{split}
     P^{(v)}_{i,k|k}  =  [I -  K^{(v)}_{i,k} H_{ii,k}] P_{i,k|k-1}^{(v)}, \nonumber
\label{eqn:PHDupdatepki}
\end{split}
\end{equation}
\noindent
\begin{equation}
\begin{split}
     K^{(v)}_{i,k}  =  P_{i,k|k-1}^{(v)} H^T_{ii,k} [ H_{ii,k}P_{i,k|k-1}^{(v)} H^T_{ii,k} + R_{ii,k}]^{-1}, \nonumber
\label{eqn:PHDupdateKki}
\end{split}
\end{equation}
\noindent $c_{s_{i,k}}(z)$ is given in Eq.~(\ref{eqn:Clutterscenei}).
Therefore, all that is left is to formulate the implementation scheme for $c_{t_{i,k}}(z)$ which is given in Eq.~(\ref{eqn:Clutteri}) and is given again as

\begin{equation}
\begin{array} {lll}
    c_{t_{i,k}}(z) = \sum_{j \in \{1,...,N\}\setminus i} \int p_{ji,D}(y)\mathcal{D}_{j,k|k-1}(y)f_{ji,k}(z|y)dy,
\end{array}
\label{eqn:Clutteri2}
\end{equation}
\noindent where $\mathcal{D}_{j,k|k-1}(y)$ is given in Eq.~(\ref{eqn:PHDpredictionki}), $f_{ji,k}(z|y)$ is given in Eq.~(\ref{eqn:linearObservation1}) and $p_{ji,D}(y)$ is assumed constant.
Since $w_{j,k|k-1}^{(v)}$ is independent of the integrable variable $y$, Eq.~(\ref{eqn:Clutteri2}) becomes

\begin{equation}
\begin{array} {lll}
c_{t_{i,k}}(z) = & \sum_{j \in \{1,...,N\}\setminus i} \sum_{v = 1}^{V_{j,k|k-1}} p_{ji,D} w_{j,k|k-1}^{(v)} \int \mathcal{N}(y; \\& m_{j,k|k-1}^{(v)}, P_{j,k|k-1}^{(v)})\mathcal{N}(z;H_{ji,k} y, R_{ji,k})dy,
\label{eqn:Clutteri3}
\end{array}
\end{equation}
\noindent This can be simplified further using the following equality given that $P_1$ and $P_2$ are positive definite

\begin{equation}
\int \mathcal{N} (y; m_1 \zeta, P_1)\mathcal{N}(\zeta; m_2, P_2)d\zeta =  \mathcal{N}(y; m_1 m_2, P_1 + m_1 P_2 m_2^T).
\label{eqn:Lemma1}
\end{equation}
\noindent Therefore,~(\ref{eqn:Clutteri3}) becomes,

\begin{equation}
\begin{array} {lll}
c_{t_{i,k}}(z) =& \sum_{j \in \{1,...,N\}\setminus i} \sum_{v = 1}^{V_{j,k|k-1}}  p_{ji,D}  w_{j,k|k-1}^{(v)} \mathcal{N}(z; \\& H_{ji,k} m_{j,k|k-1}^{(v)}, R_{ji,k} + H_{ji,k} P_{j,k|k-1}^{(v)}H_{ji,k}^T),
\label{eqn:Clutteri4}
\end{array}
\end{equation}
\noindent where $i \in \{1,..., N\}$.

\begin{algorithm}
\caption{Pseudocode for the N-type GM-PHD filter}\label{alg:tri-GMPHD}
\begin{algorithmic}[1]
\State \textbf{given} {\{$w_{i,k-1}^{(v)}, m_{i,k-1}^{(v)}, P_{i,k-1}^{(v)}\}_{v=1}^{V_{i,k-1}}$ for target type $i \in \{1, ..., N\}$, and the measurement set $Z_{j,k}$ for $j \in \{1, ..., N\}$}
\State \textbf{step 1.} {(prediction for birth targets)}
\For {$i = 1, ..., N$} \Comment{for all target type $i$}
\State $e_i = 0$
\For{$u = 1, ..., V_{\gamma_i,k}$}
\State $e_i := e_i + 1$
\State $w_{i,k|k-1}^{(e_i)} = w_{i,\gamma,k}^{(u)}$
\State $m_{i,k|k-1}^{(e_i)} = m_{i,\gamma,k}^{(u)}$
\State $P_{i,k|k-1}^{(e_i)} = P_{i,\gamma,k}^{(u)}$
\EndFor
\EndFor
\State \textbf{step 2.} {(prediction for existing targets)}
\For {$i = 1, ..., N$}  \Comment{for all target type $i$}
\For {$u = 1, ..., V_{i,k-1}$}
\State $e_i := e_i + 1$
\State $w_{i,k|k-1}^{(e_i)} = p_{i,S,k}w_{i,k-1}^{(u)}$
\State $m_{i,k|k-1}^{(e_i)} = F_{i,k-1}m_{i,k-1}^{(u)}$
\State $P_{i,k|k-1}^{(e_i)} = Q_{i,k-1} + F_{i,k-1} P_{i,k-1}^{(u)} F_{i,k-1}^T$
\EndFor
\EndFor
\State $V_{i,k|k-1} = e_i$
\State \textbf{step 3.} {(Construction of PHD update components)}
\For {$i = 1, ..., N$}  \Comment{for all target type $i$}
\For {$u = 1, ..., V_{i,k|k-1}$}
\State $\eta_{i,k|k-1}^{(u)} = H_{ii,k} m_{i,k|k-1}^{(u)}$
\State $S_{i,k}^{(u)} = R_{ii,k} + H_{ii,k}P_{i,k|k-1}^{(u)} H^T_{ii,k}$
\State $K^{(u)}_{i,k}  =  P_{i,k|k-1}^{(u)} H^T_{ii,k} [ S_{i,k}^{(u)} ]^{-1}$
\State $P^{(u)}_{i,k|k}  =  [I -  K^{(u)}_{i,k} H_{ii,k}] P_{i,k|k-1}^{(u)}$
\EndFor
\EndFor
\State \textbf{step 4.} {(Update)}
\For {$i = 1, ..., N$}  \Comment{for all target type $i$}
\For {$u = 1, ..., V_{i,k|k-1}$}
\State $w_{i,k}^{(u)} = (1 - p_{ii,D,k})w_{i,k|k-1}^{(u)}$
\State $m_{i,k}^{(u)} = m_{i,k|k-1}^{(u)}$
\State $P_{i,k}^{(u)} = P_{i,k|k-1}^{(u)}$
\EndFor
\State $l_i := 0$
\For{each $z \in Z_{j,k}$}
\State $l_i := l_i + 1$
\For {$u = 1, ..., V_{i,k|k-1}$}
\State $w_{i,k}^{(l_i V_{i,k|k-1} + u)} =~ p_{ii,D,k}w_{i,k|k-1}^{(u)}\mathcal{N}(z;
\myindent{4.2} \eta_{i,k|k-1}^{(u)}, S_{i,k}^{(u)}) $
\State $m_{i,k}^{(l_i V_{i,k|k-1} + u)} =~ m_{i,k|k-1}^{(u)} + K^{(u)}_{i,k} (z -
\myindent{4.3} \eta_{i,k|k-1}^{(u)})$
\State $P_{i,k}^{(l_i V_{i,k|k-1} + u)} = P^{(u)}_{i,k|k}$
\EndFor
\algstore{myalg}
\end{algorithmic}
\end{algorithm}

\begin{algorithm}
\begin{algorithmic} [1]                   
\algrestore{myalg}
\For {$u = 1, ...., V_{i,k|k-1}$}
\State $c_{s_{i,k}}(z) = \lambda_{c_i} A c_i(z)$
\State $c_{t_{i,k}}(z) = \sum_{j \in \{1,...,N\}\setminus i} \sum_{e = 1}^{V_{j,k|k-1}}  p_{ji,D}  w_{j,k|k-1}^{(e)}
\myindent{1} \mathcal{N}(z; H_{ji,k} m_{j,k|k-1}^{(e)}, R_{ji,k} + H_{ji,k} P_{j,k|k-1}^{(e)}H_{ji,k}^T)$  \label{TriGMPHDconfusion}
\State $c_{i,k}(z) = c_{s_{i,k}}(z) + c_{t_{i,k}}(z)$
\State $w_{i,k,N} = \sum_{e=1}^{V_{i,k|k-1}} w_{i,k}^{(l_i V_{i,k|k-1} + e)}$
\State $w_{i,k}^{(l_i V_{i,k|k-1} + u)} = \frac{w_{i,k}^{(l_i V_{i,k|k-1} + u)}}{c_{i,k}(z) + w_{i,k,N}}$
\EndFor
\EndFor
\State $V_{i,k} = l_i V_{i,k|k-1} + V_{i,k|k-1}$
\EndFor
\State \textbf{output} {\{$w_{i,k}^{(v)}, m_{i,k}^{(v)}, P_{i,k}^{(v)}\}_{v=1}^{V_{i,k}}$}
\end{algorithmic}
\label{alg:the_alg}
\end{algorithm}


The key steps of the N-type GM-PHD filter are summarised in Algorithms~\ref{alg:tri-GMPHD} and \ref{alg:tri-GMPHDprune}.
These are expressed in terms of frames $k$ and $k-1$; for the first frame, $k=1$, of a sequence there is only detection and target birth, but no prediction and update for existing targets.
For subsequent frames, we have chosen measurement driven target birth, rather than a random or a-priori birth model, inspired by but not identical to ~\cite{RisClaVoVo12}.
Maggio et al.~\cite{MagTajCav08} also assume that targets are born in a limited volume around measurements.
The advantage of random birth is in the potential detection of weak target signatures, but in these examples the presence of a human should, in general, generate a strong probability of detection provided the target is in view.
This is borne out by experiments and parameter setting in Section \ref{Sec:ExperimentalResults}.
A further disadvantage of random birth is the increased complexity of processing a large number of incorrect targets.
For targets moving in video sequences there is no spawn process, but occlusions do result anywhere in the field of view, and may be caused either by other targets or other obstacles.
Re-emerging targets are detected and constitute births, are not spawned because they may be occluded by obstacles other than targets, and have no a-priori location.

The prediction and update, steps 2 to 4, follow the standard procedures for the GM-PHD filter~\cite{VoMa06} but are extended to take into account the $N$ detection processes and the subsequent confusion between detections.
In the proposed algorithm, birth and prediction both precede the construction and update of the PHD components, so the total number at the conclusion of step 4 is the sum of the persistent and birthed components. The number of Gaussian components in the posterior intensities may increase without bound as time progresses, particularly as a birth at this stage may be due to an existing target that has moved from the previous frame and then is re-detected in the current frame. Therefore, it is necessary to prune weak and duplicated components in Algorithm~\ref{alg:tri-GMPHDprune}.
First, weak components with weight $w_{i,k}^{(v)} < T = 10^{-5}$ are pruned.
Further, Gaussian components with Mahalanobis distance less than $U = 4$ pixels from each other are merged. These pruned and merged Gaussian components, output of Algorithm~\ref{alg:tri-GMPHDprune}, are predicted as existing targets in the next iteration.
Finally, Gaussian components of the posterior intensity, output of Algorithm~~\ref{alg:tri-GMPHD}, with means corresponding to weights greater than 0.5 as a threshold are selected as multi-target state estimates.

\begin{algorithm}
\caption{Pruning and merging for the N-type GM-PHD filter}\label{alg:tri-GMPHDprune}
\begin{algorithmic}[1]
\State \textbf{given} {\{$w_{i,k}^{(v)}, m_{i,k}^{(v)}, P_{i,k}^{(v)}\}_{v=1}^{V_{i,k}}$ for target type $i \in \{1, ..., N\}$, a pruning weight threshold T, and a merging distance threshold U.}
\For {$i = 1, ..., N$} \Comment{for all target type $i$}
\State  {Set $\ell_i = 0$, and $I_i = \{ v = 1, ..., V_{i,k}|w_{i,k}^{(v)} > T$ \}}
\State \textbf{repeat}
\State $\ell_i := \ell_i + 1$
\State $u := \arg\max_{v \in I_i} w_{i,k}^{(v)}$
\State $L_i := \Big\{v \in I_i \Big| (m_{i,k}^{(v)} - m_{i,k}^{(u)})^T (P_{i,k}^{(v)})^{-1} (m_{i,k}^{(v)} - m_{i,k}^{(u)}) \leq U \Big \}$
\State $\tilde{w}_{i,k}^{(\ell_i)} = \sum_{v \in L_i} w_{i,k}^{(v)}$
\State $\tilde{m}_{i,k}^{(\ell_i)} = \frac{1}{\tilde{w}_{i,k}^{(\ell_i)}} \sum_{v \in L_i} w_{i,k}^{(v)} m_{i,k}^{(v)}$
\State $\tilde{P}_{i,k}^{(\ell_i)} = \frac{1}{\tilde{w}_{i,k}^{(\ell_i)}} \sum_{v \in L_i} w_{i,k}^{(v)} (P_{i,k}^{(v)} + (\tilde{m}_{i,k}^{(\ell_i)} - m_{i,k}^{(v)})(\tilde{m}_{i,k}^{(\ell_i)} - m_{i,k}^{(v)})^T)$
\State $I_i := I_i \setminus L_i$
\State \textbf{until} {$I_i = \emptyset$}
\EndFor
\State \textbf{output} {\{$\tilde{w}_{i,k}^{(v)}, \tilde{m}_{i,k}^{(v)}, \tilde{P}_{i,k}^{(v)}\}_{v=1}^{\ell_i}$} as pruned and merged Gaussian components for target type $i$.

\end{algorithmic}
\end{algorithm}

%

\section{Experimental Results} \label{Sec:ExperimentalResults}

We apply the N-type GM-PHD filter to video sequences by integrating the object detectors' information such as the probabilities of detections for each target type and the confusion detection probabilities among target types at a specific background clutter rate. Accordingly, we consider different scenarios as follows.

\subsection{Multiple Target, Multiple Type Tracking using a Tri-GM-PHD Filter} \label{Sec:threeTypes}

In this part, we consider tracking of football teams and a referee in the same scene handling their confusions using a tri-GM-PHD filter ($N=3$).
As the three types of target are all sub-categories of a human type, the targets has the same aspect ratio in this example, and indeed color is the primary discriminator in the detection process.

\subsubsection{Object Detection, Training and Evaluation} \label{Sec:ObjectDetection}

The RFS methodology post-processes a set of detections with parameters defining the probabilities of detection and clutter (false alarms) which arise from trials using the detector.
For the tri-PHD filter, we also need parameters for confusion.
We employ the existing, state-of-the-art, Aggregated Channel Features (ACF) pedestrian detector~\cite{DolAppPerBel14} adapted to our data set due to its computational efficiency and ease of use.
This uses three different kinds of features in 10 channels: normalized gradient magnitude (1 channel), histograms of oriented gradients (6 channels), and LUV color (3 channels).
It is applied to detect the actors (football teams and a referee) using a sliding window
at multiple scales.
The Adaboost classifier~\cite{AppFucDolPer13} is used to learn and classify the feature vectors acquired by the ACF detector.

For training, evaluation and parameter setting we use the VS-PETS'2003 football video data\footnote{http://www.cvg.reading.ac.uk/slides/pets.html}.
This consists of 2500 frames which have players from the red and white teams and the referee.
We trained 3 separate detectors for each target type (red, white, referee).
We used every 10'th frame, i.e. 240 frames taken from the last 2400 frames, including 2000 positive samples for each footballer type, 240 samples for the referee,
and 5000 randomly selected negative samples.
This captures the appearance variation of players due to articulated motion.
The correct player type or referee positions and windows were labeled manually for training as positive samples.
The first 100 frames (video) are used to evaluate and test the tri-GM-PHD filter in comparison with repeated detection and three separate GM-PHD filters in section~\ref{Sec:trackingResults}.

\begin{figure*} [!htb] 
  \begin{center}
   \subfloat[\scriptsize{ROC for red team detector on red football instances}]
  {\label{fig:ROCgcnnPs16} \includegraphics[height=0.23\textwidth]{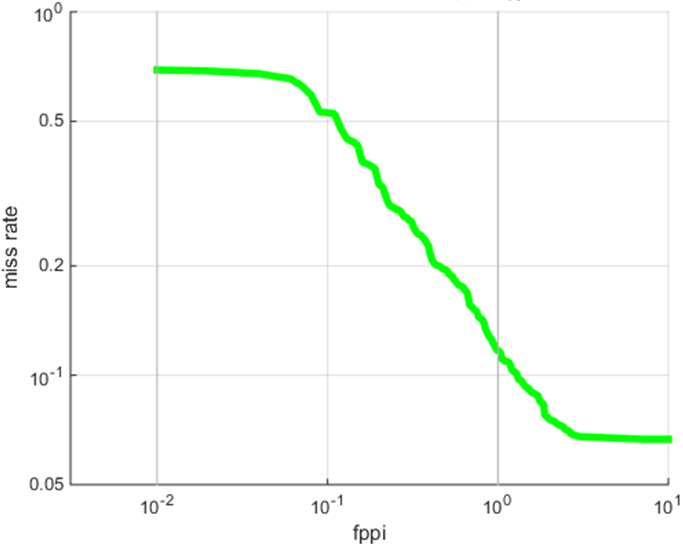}} 
   	\subfloat[\scriptsize{ROC for white team detector on white football instances}]
  {\label{fig:ROCgcnnPs16} \includegraphics[height=0.23\textwidth]{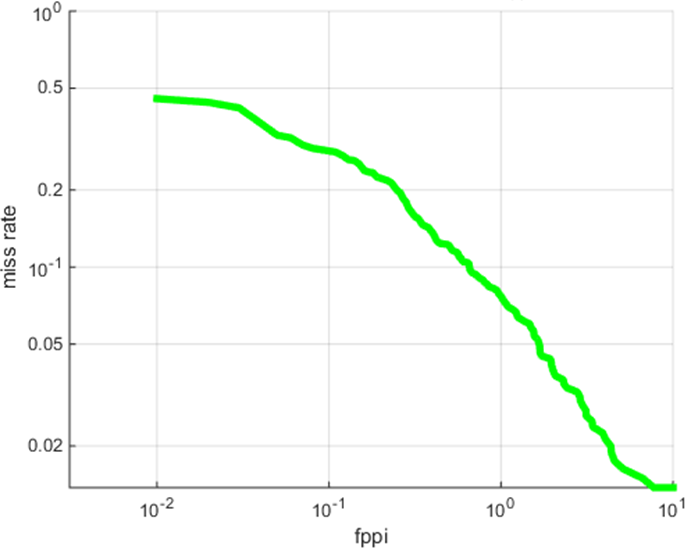}} 
	  \subfloat[\scriptsize{ROC for referee detector on referee instances}]
  {\label{fig:ROCgcnnPs16} \includegraphics[height=0.23\textwidth]{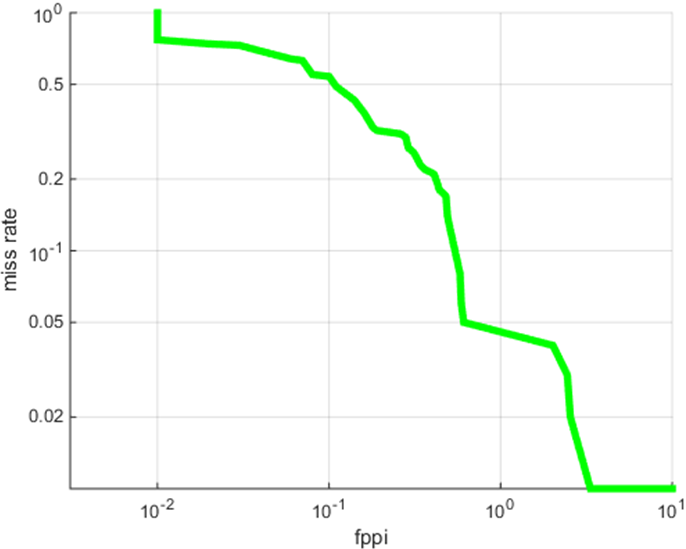}} 
	\caption{\small{Extracting detection probabilities for three target types (red, white and referee) from ROCs of 3 detectors: red team detector, white team detector and referee detector when tested on red team instances, white team instances and referee instances, respectively.}} 
  \label{fig:ROCDetectors}
  \end{center}
\end{figure*}

The RFS methodology assumes point detections and a Gaussian error distribution on accuracy of location.
However, humans in a video sequence are extended targets and the ACF detector has a bounding box that encloses the target.
Therefore, overlapping detections are merged using a greedy non-maximum suppression (NMS) overlap threshold (intersection over union of two detections) of 0.05 (we made the overlap threshold very tight to ignore multiple bounding boxes on the same object).
However, when evaluating the detectors, an overlap threshold (intersection over union of detection and ground truth bounding boxes) of 0.5 is used to identify true positives vs false positives.
The receiver operating characteristic (ROC) curves for each of the detectors are given in
Fig.~\ref{fig:ROCDetectors}.

\noindent

For the tri-GM-PHD strategy, we must set the thresholds on detection from the
ROC curves in Fig.~\ref{fig:ROCDetectors}, taking into account the probabilities of confusion that arise from the corresponding ROC curves (not shown) of each detector applied to targets of a confusing type.
From our own simulations and the published literature, e.g. \cite{VoMa06,RisClaVoVo12}, we know that the RFS methodology is most effective when applied with a high probability of detection, albeit with a higher clutter rate, and in our case a higher confusion rate. Obviously, for a target detection to be useful, the probability of true detection must be higher than the probability of confusion.
Therefore, from Fig.~\ref{fig:ROCDetectors}, we standardise a clutter rate of $10$ false positive per image (fppi), which gives probabilities of detection of $0.93$ ($p_{11}$), $0.99$ ($p_{22}$) and $0.99$ ($p_{33}$) for red, white and referee, respectively.
With these values, the corresponding confusion parameters are $0.24$ (white footballer detected as red i.e. applying red detector on white footballer instances, $p_{12}$), $0.5$ (referee detected as red, $p_{13}$), $0.24$ (red as white, $p_{21}$), $0.18$ (referee as white, $p_{23}$), $0.19$ (red as referee, $p_{31}$) and $0.17$ (white as referee, $p_{32}$). More clearly, the probabilities of confusion are obtained from the corresponding ROC curves of each detector applied to targets of a confusing type at fppi of 10. For instance, $p_{12}$ is obtained from the ROC curve of red detector applied to white footballer instances at fppi of 10.

\subsubsection{Data Association}\label{Sec:DataAssociation}

The tri-GM-PHD filter handles sensor noise, clutter and distinguishes between true and false targets of each type.
However, this does not distinguish between two different targets of the same type, so
an additional step can be applied if we wish to identify different
targets of the same type between consecutive frames.
Although not part of the tri-GM-PHD strategy, this is commonly required so we include results from this
post-labeling process for completeness in section \ref{Sec:trackingResults}.
It does not affect our error metrics but is a post-process to label individuals from frame to frame.
For data association, the Euclidean distance between each previous filtered centroid (track) and the current filtered centroids is computed and we compute an assignment which minimizes the total cost returning assigned tracks to current filtered outputs.
This assignment problem represented by the cost matrix is solved using Munkres's variant of the Hungarian algorithm~\cite{FraJea71}.

This also returns the unassigned tracks and unassigned current filtered results.
The unassigned tracks are deleted and the unassigned current filtered outputs
create new tracks if the targets are not created earlier.
If some targets are miss-detected and incorrectly labeled, labels are uniquely re-assigned by re-identifying them using the approach in~\cite{EjaMic15}.


\subsubsection{Tracking Results}\label{Sec:trackingResults}

Referring to Eq.~(\ref{eqn:stateSet1}), our state vector includes the centroid positions, velocities, and the width and height of the bounding boxes,  i.e. $x_k = [p_{cx,xk}, p_{cy,xk}, \dot{p}_{x,xk}, \dot{p}_{y,xk}, w_{xk}, h_{xk}]^T$.
Similarly, the measurement is the noisy version of the target area in the image plane approximated with a $w$ x $h$ rectangle centered at $(p_{cx,zk}, p_{cy,zk})$ i.e. $z_k = [p_{cx,zk}, p_{cy,zk}, w_{zk}, h_{zk}]^T$.

As stated above, the detection and confusion probabilities are set by experimental evaluation of the ACF detection processes.
Additional parameters are set from simulation and previous experience.
For each target type, we set survival probabilities $p_{1,S} = p_{2,S} = p_{3,S} = 0.99 $, and we assume the linear Gaussian dynamic model of Eq.~(\ref{eqn:linearState1}) with matrices taking into account the box width and height at the given scale.

\[ F_{i,k-1} = \left[ \begin{array}{ccc}
           I_2 & \Delta I_2 & 0_2 \\
           0_2 & I_2 & 0_2 \\
           0_2 & 0_2 & I_2
           \end{array} \right], \]
													
\begin{equation} Q_{i, k-1} = \sigma_{v_i}^2 \left [ \begin{array}{ccc}
    \frac{\Delta^4}{4}I_2 & \frac{\Delta^3}{2}I_2  & 0_2\\
    \frac{\Delta^3}{2}I_2 & \Delta^2 I_2 & 0_2 \\
    0_2 & 0_2 & \Delta^2 I_2
    \end{array} \right],
\label{eqn:PHDstateTransitionMatrixVideo}
\end{equation}
\noindent
where $I_n$ and $0_n$ denote the \textit{n} x \textit{n} identity and zero matrices, respectively and $\Delta$ is the sampling period defined by the time between frames (we use 1 second).  $\sigma_{v_i} = 5$ pixels$/s^2$ are the standard deviations of the process noise for target type $i$ where $i \in \{1, 2, 3\}$ i.e. type 1 (red football team), target type 2 (white football team) and target type 3 (a referee).

Similarly, the measurement follows the observation model of Eq.~(\ref{eqn:linearObservation1}) with matrices taking into account the box width and height,

\[H_{ii,k} = H_{ji,k} = \left[ \begin{array}{ccc}
           I_2 & 0_2 & 0_2 \\
           0_2 & 0_2 & I_2
           \end{array} \right], \]

\[R_{ii, k} = \sigma_{r_{ii}}^2 \left [ \begin{array}{cc}
     I_2 & 0_2 \\
     0_2 & I_2
    \end{array} \right], \]

\begin{equation} R_{ji, k} = \sigma_{r_{ji}}^2 \left [ \begin{array}{cc}
     I_2 & 0_2 \\
     0_2 & I_2
    \end{array} \right],
\label{eqn:eqn:PHDobservationMatrixVideo}
\end{equation}
\noindent where $i \in \{1, 2, 3\}$, $j \in \{1, 2, 3\}$, and $\sigma_{r_{ii}}$ and $\sigma_{r_{ji}}$ are the measurement standard deviations taken from the distribution of distance errors of the centroids from ground truth in the evaluation of the detection process, effectively 6 pixels.

Accordingly, in our approach, positive detections specify the possible birth locations with the initial covariance given in Eq.~(\ref{eqn:PHDbirthCovariance}).
The current measurement and zero initial velocity are used as a mean of the Gaussian distribution using a pre-determined initial covariance for birthing of targets, i.e. new targets are born in the region of the state space for which the likelihood will have high values. Precisely, the birthing of targets is completely automatic using the very recent measurements obtained from object detectors. Very small initial weight (e.g. $10^{-4}$) is assigned to the Gaussian components for new births as this is effective for high clutter rates. This is basically equivalent to the average number of appearing (birth) targets per frame ($n_b$) divided uniformly across the frame resolution ($A$).

\begin{equation}
 P_{i,\gamma,k} = diag([100, 100, 25, 25, 20, 20]).
\label{eqn:PHDbirthCovariance}
\end{equation}
\noindent where $i \in \{1, 2, 3\}$.

\begin{figure*} [!htb] 
  \begin{center}
   \subfloat[\scriptsize{Detections, frame 25}]
  {\label{fig:DetectionFrame25} \includegraphics[height=0.23\textwidth]{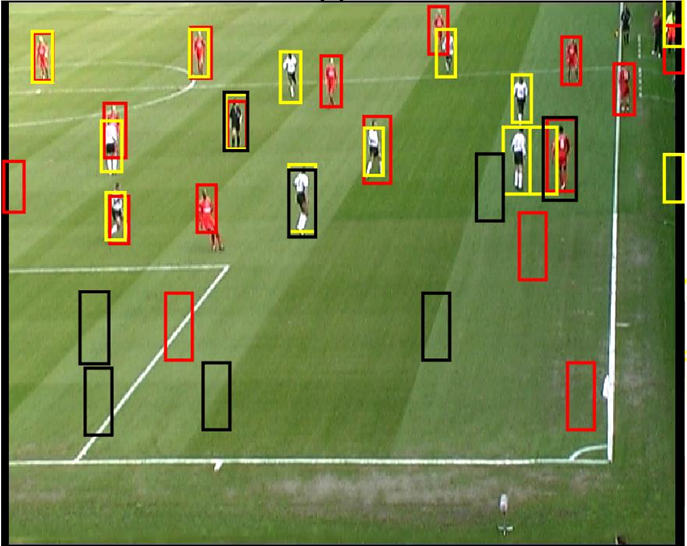}} 
   \subfloat[\scriptsize{Three independent GM-PHD trackers, frame 25 }]
  {\label{fig:ThreePHDsFrame25} \includegraphics[height=0.23\textwidth]{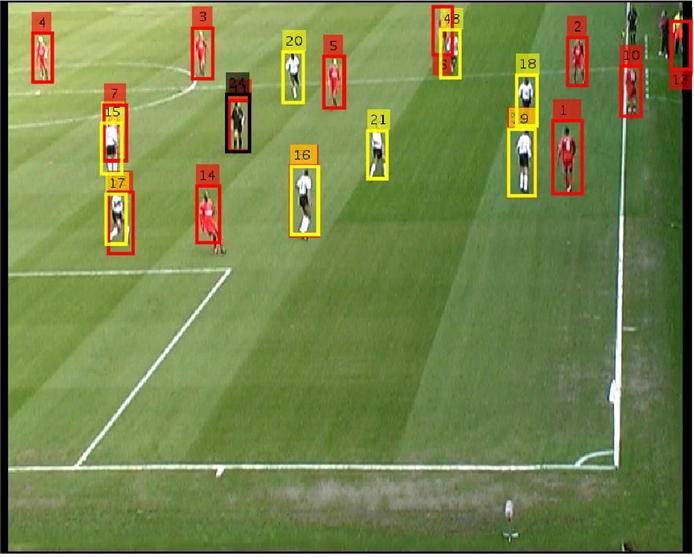}}
  \subfloat[\scriptsize{Tri-GM-PHD Tracker, frame 25}]
  {\label{fig:TriPHDFrame25} \includegraphics[height=0.23\textwidth]{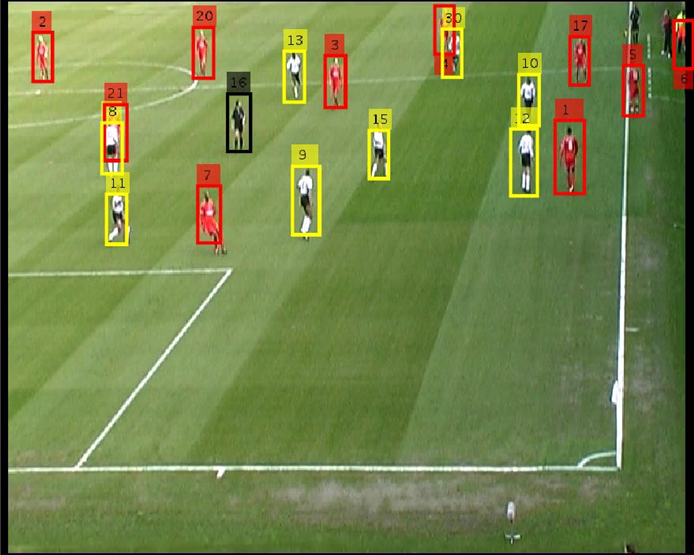}} 
  \end{center}
   \caption{\small{Results of detections, three independent GM-PHD trackers and tri-GM-PHD tracker, respectively, for frame 25.}}
  \label{fig:TrackingFrame25}
\end{figure*}

\begin{figure*} [!htb] 
  \begin{center}
   \subfloat[\scriptsize{Detections, frame 57}]
  {\label{fig:DetectionFrame57} \includegraphics[height=0.23\textwidth]{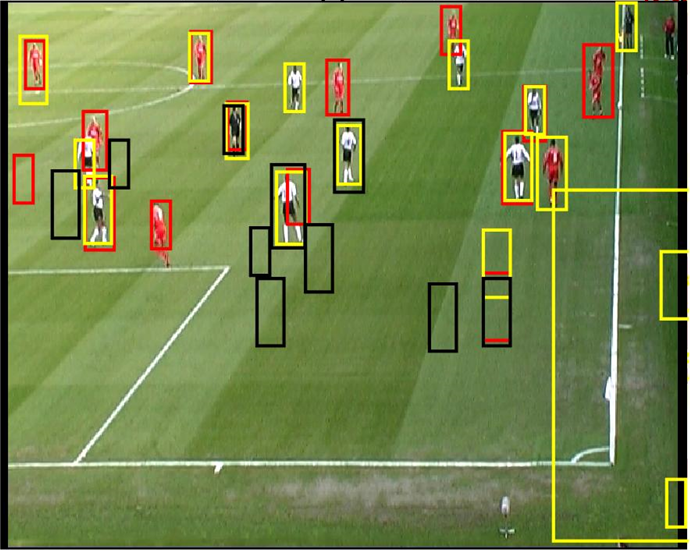}} 
   \subfloat[\scriptsize{Three independent GM-PHD trackers, frame 57}]
  {\label{fig:ThreePHDsFrame57} \includegraphics[height=0.23\textwidth]{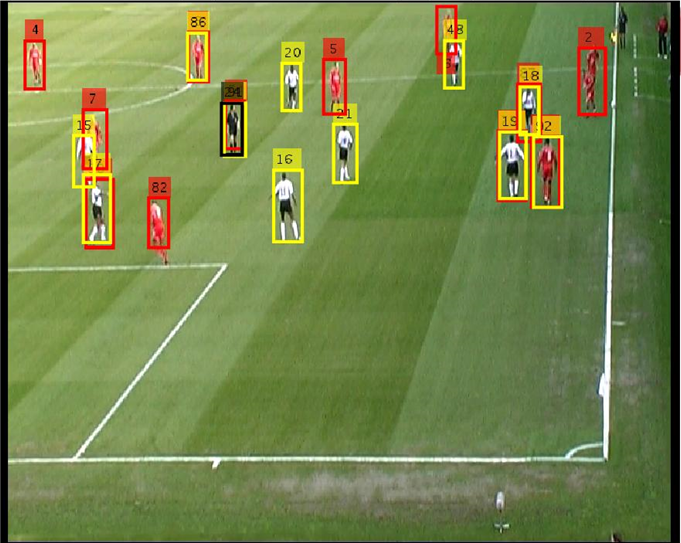}}
  \subfloat[\scriptsize{Tri-GM-PHD Tracker, frame 57}]
  {\label{fig:TriPHDFrame57} \includegraphics[height=0.23\textwidth]{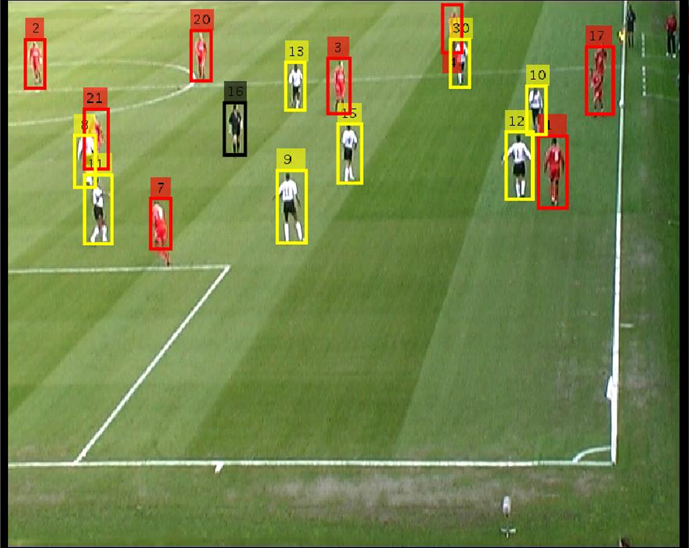}} 
  \end{center}
   \caption{\small{Results of detections, three independent GM-PHD trackers and tri-GM-PHD tracker, respectively, for frame 57.}}
  \label{fig:TrackingFrame57}
\end{figure*}

\begin{figure*} [!htb] 
  \begin{center}
   \subfloat[\scriptsize{Detections, frame 73}]
  {\label{fig:DetectionFrame73} \includegraphics[height=0.23\textwidth]{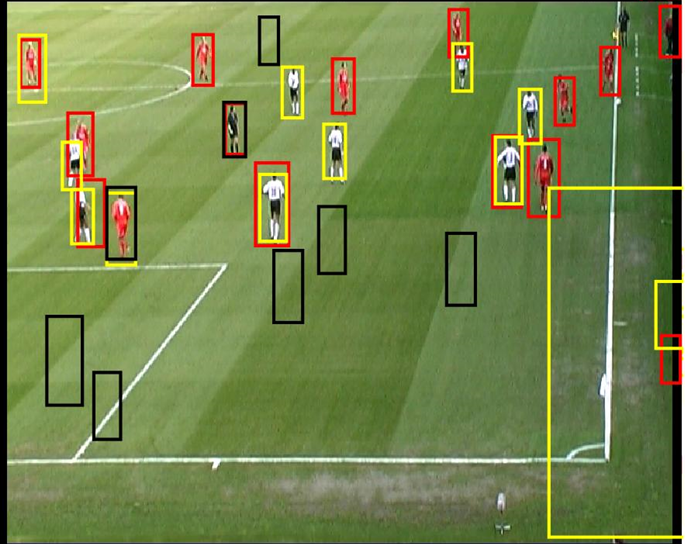}} 
   \subfloat[\scriptsize{Three independent GM-PHD trackers, frame 73}]
  {\label{fig:ThreePHDsFrame73} \includegraphics[height=0.23\textwidth]{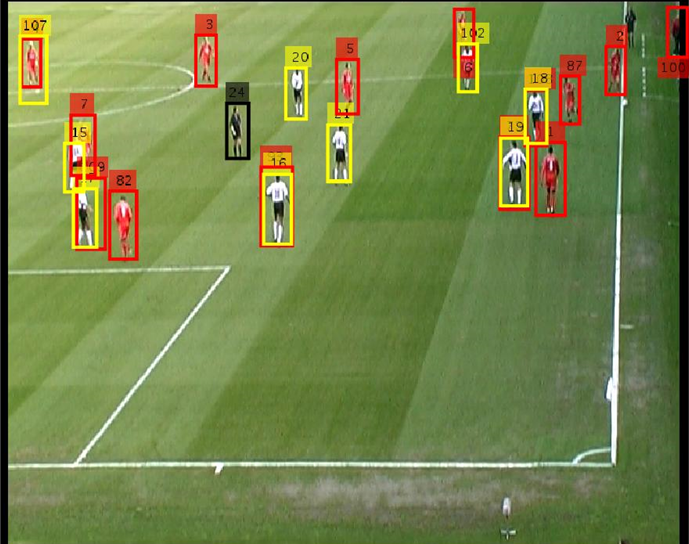}}
  \subfloat[\scriptsize{Tri-GM-PHD Tracker, frame 73}]
  {\label{fig:TriPHDFrame73} \includegraphics[height=0.23\textwidth]{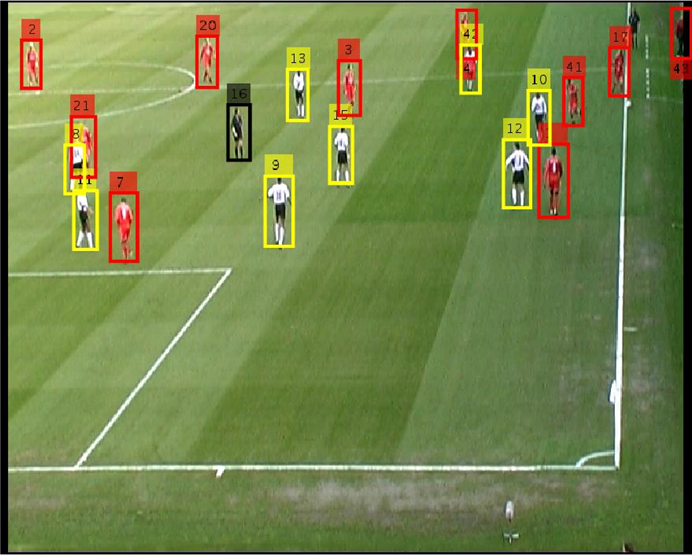}} 
  \end{center}
   \caption{\small{Results of detections, three independent GM-PHD trackers and tri-GM-PHD tracker, respectively, for frame 73.}}
  \label{fig:TrackingFrame73}
\end{figure*}

We evaluate the tracking methodology of the tri-GM-PHD tracker in comparison with first, repeated independent detection on each frame, and second, with three independent GM-PHD trackers. Using the football video sequence, the examples shown in Fig.~\ref{fig:TrackingFrame25}, Fig.~\ref{fig:TrackingFrame57} and Fig.~\ref{fig:TrackingFrame73} are for repeated detection (no tracking), three independent GM-PHD trackers, and the tri-GM-PHD tracker for frames 25, 57 and 73, respectively.
Hence, Fig.~\ref{fig:DetectionFrame57} designates detections in which the red footballers, white footballers and the referee are detected both correctly and incorrectly, i.e. one object may be detected by many detectors.
In this example the referee is detected 3 times: by the red team detector (red), by the white team detector (yellow) and the referee detector (black).
Moreover, there are many background false positives (clutter) in the scene that arise from our
choice to set the detection probability high at the expense of higher clutter as this is the
detection scenario that is favored by the PHD process.
Using the three independent GM-PHD trackers to effectively eliminate false positives, confused detections are not resolved as shown in Fig.~\ref{fig:ThreePHDsFrame57}.
However, our proposed tri-GM-PHD tracker effectively eliminates the false positives as well as confused detections as shown in Fig.~\ref{fig:TriPHDFrame57}.

\begin{figure*} [!htb] 
  \begin{center}
   \subfloat[\scriptsize{Cardinality}]
  {\label{fig:VisCardinality} \includegraphics[height=0.35\textwidth]{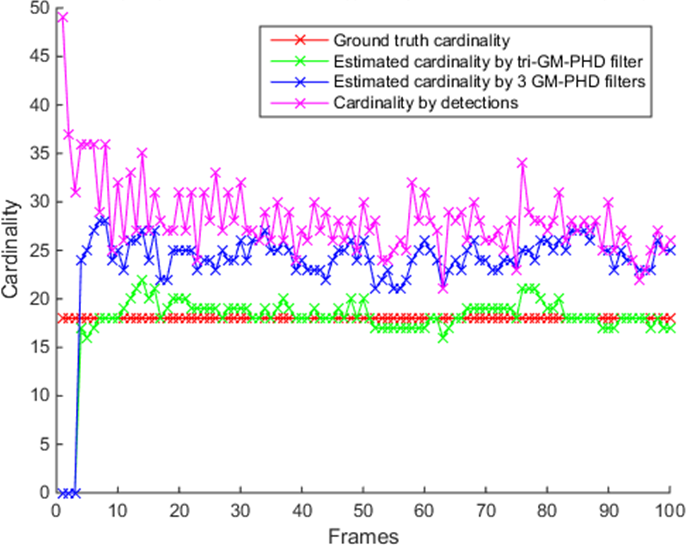}} 
   \subfloat[\scriptsize{OSPA error}]
  {\label{fig:VisOSPAerror} \includegraphics[height=0.35\textwidth]{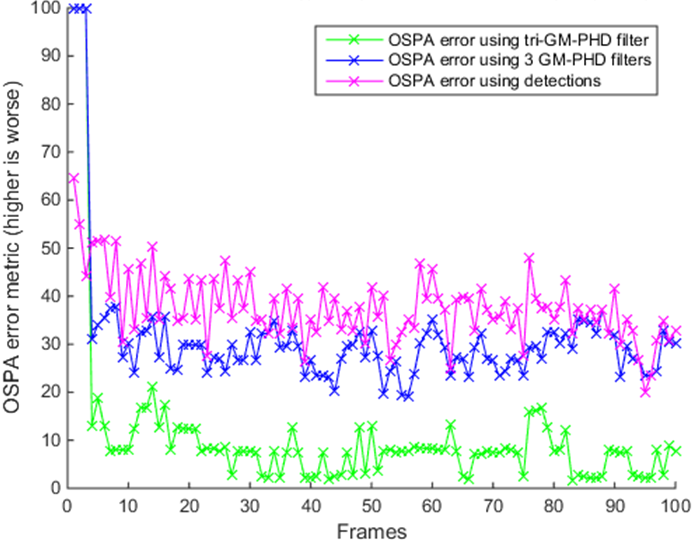}}
  \end{center}
   \caption{\small{Cardinality and OSPA error: Ground truth (red for cardinality only), tri-GM-PHD filter (green), three independent GM-PHD filters (blue), detections (magenta).}}  
  \label{fig:OSPA-CardinalityComparisonVis}
\end{figure*}

The tri-GM-PHD filter is evaluated quantitatively for the whole test sequence and compared with three independent GM-PHD filters and repeated detection using cardinality, the OSPA metric~\cite{SchVoVo08} (using order $p = 1$ and cutoff $c = 100$), discrimination rate and time taken. We use the OSPA metric which is designed for evaluating RFS-based filters rather than multi-object tracking accuracy (MOTA)~\cite{BerSti08} which is widely used for evaluating other traditional multi-target tracking algorithms~\cite{YooLeeYan16,Choi15}. Our algorithm is developed not only for tracking but also for discriminating different target types overcoming their confusions unlike algorithms such as~\cite{YooLeeYan16,Choi15}. Therefore, the OSPA is the right evaluation metric to compare our approach with repeated raw detection and three independent GM-PHD trackers. The computational figures arise from experiments on a i5 2.50 GHz core processor with 6 GB RAM laptop using MATLAB and we acknowledge that these are not definitive and give a rough guide only to implementation costs.
Though labeling of the targets using Munkres's variant of the Hungarian assignment algorithm works well as shown in Figs.~\ref{fig:TriPHDFrame25}, \ref{fig:TriPHDFrame57} and \ref{fig:TriPHDFrame73}, we did not include this in our evaluation as it is not part of the quantitative comparison of the filtering and type labeling of either the detection or distinct GM-PHD filters.
We present the cardinality and OSPA error plots in Fig.~\ref{fig:VisCardinality} and Fig.~\ref{fig:VisOSPAerror} respectively, in red for ground truth (cardinality), green for the tri-GM-PHD filter, blue for the three independent GM-PHD filters and magenta for repeated detection.
As summarised in Table~\ref{tbl:OSPAerror} the average absolute cardinality
error using detection only is 10.22, reduced to 5.76 using the standard GM-PHD filters
and to 0.11 using the tri-GM-PHD filter.
The overall frame-averaged value of OSPA error for the tri-GM-PHD filter is 10.59 pixels, compared to three independent GM-PHD filters of 30.86 pixels, and repeated detections of 37.61 pixels.
The proposed approach reduces the cardinality and OSPA errors by a large margin over three independent GM-PHD filters and repeated detection, although this has more computational cost as also shown in Table~\ref{tbl:OSPAerror}.

Independent GM-PHD trackers do not take confusion into account, so treat such confusion as 'background' clutter; the problem is that such confused detections are not likely to be accurately modeled by random detections distributed uniformly in space as is commonly the case.
Our approach can effectively discriminate true positives from clutter, while eliminating confused detections with a discrimination rate of 99.20\%.
The mis-discrimination rate of 0.80\% occurs primarily during the initial frames (e.g. the first 7 frames) until the prediction-update process stabilises and the true detections are confirmed by the motion between adjacent frames.

\noindent
\noindent
\noindent
\noindent
\begin{table*}[!htb]
\begin{center}
\begin{tabular}{|l|c|c|c|r|}
\hline
Method & cardinality error & OSPA error & time taken & discrimination rate \\
\hline\hline
Detections & 10.22 & 37.61 pixels & 0.59 seconds/frame & 0\% \\
3 GM-PHDs  &  5.76 & 30.86 pixels & 0.80 seconds/frame & 0\% \\
Tri-GM-PHD &  0.11 & 10.59 pixels & 3.00 seconds/frame & 99.20\% \\
\hline
\end{tabular}
\end{center}
\caption{\small{Cardinality and OSPA errors, time taken and discrimination rate at the extracted detection probabilities for tri-GM-PHD filter, three independent GM-PHD filters and Detections.}}
\label{tbl:OSPAerror}
\end{table*}
\noindent
\begin{figure*} [!htb] 
  \begin{center}
   \subfloat[\scriptsize{Frame 193}]
  {\label{fig:Frame193} \includegraphics[height=4.3cm,width=8cm]{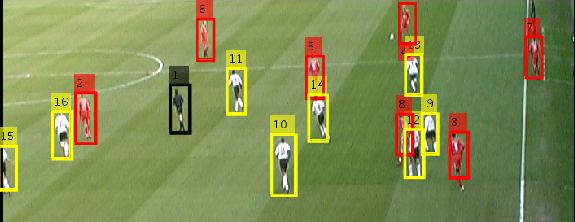}} 
   \subfloat[\scriptsize{Tracking both teams and referee, frame 293}]
	{\label{fig:Frame293} \includegraphics[height=4.3cm,width=8cm]{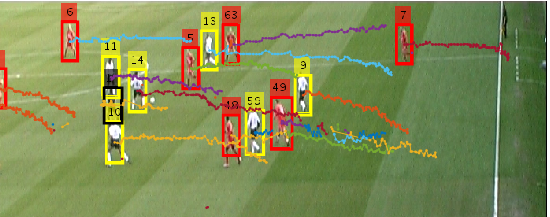}}
  \end{center}
   \caption{\small{Tracking the red and white teams, and referee from frame 193 to frame 293}}
  \label{fig:Tracking193to293}
\end{figure*}
\noindent
Fig. \ref{fig:Tracking193to293} shows another example in which the individual footballers are detected, filtered, tracked and labeled for 100 frames.
The image has been cropped as the action is confined to the top half of the image, and immediately follows a throw-in as the players move away left from the touchline.
The examples also show the individual tracks and labels of the footballers and referee as small numbers over the targets.
From this sequence, we see for example that the red player number 6 and the white player number 10, and several others, are consistently tracked through the sequence.
However the labeling does occasionally make mistakes, for example red player 3 who starts near the touchline is finally labeled as red player number 49 in frame 293.
In this instance the mislabeling is due to occlusion and lack of persistence in the detection and tracking as it uses successive frames only, so that if a player disappears then re-appears after several frames he is treated as a new target.
Nevertheless, although this evaluation is not part of the Tri-GM-PHD filter, the labeling that we apply has good performance with a mean label switch error of only 0.43\%.

\subsection{Multiple Target, multiple Type Tracking using a Dual GM-PHD Filter} \label{Sec:twoTypes}

In this part, we consider tracking of pedestrians and vehicles in the same scene handling their confusions using a dual GM-PHD filter ($N=2$).
These are fundamentally different target types; in this case they can have different aspect ratios, and whereas vehicles have radically different appearance and aspect ratio depending on azimuthal view angle, this is not the case for pedestrians.

\subsubsection{Pedestrian and Vehicle Detection} \label{Sec:ObjectDetection2}

We adapted the ACF pedestrian detector~\cite{DolAppPerBel14} by considering appearance variations. Similarly, we adapted the vehicle detector in~\cite{OhnTri15} which uses the same type of features as the ACF detector considering additional geometrical features such as truncation level, occlusion level and occlusion type features in addition to 3D orientation which depends on the ground truth information. However, we only consider 3D geometric orientation from the ground truth information available in the KITTI dataset~\cite{GeiLenStiUrt2013}. Similar to~\cite{OhnTri15}, we also consider visual features to capture appearance variations due to varying orientation, truncation and occlusion degree for detecting both pedestrians and vehicles.

\textit{3D orientation}: Appearance variation due to observation angle is common when detecting vehicles in different driving settings. Accordingly, the observation angle i.e. relative orientation of the object with respect to the camera is used by considering the angle of the vector joining the camera center in 3D and an object which takes into account the ego-vehicle. This 3D geometric orientation is available in the KITTI dataset ranging from -$\pi$ (-3.14) to $\pi$ (3.14) and is quantized into L labels ( L = 5 for pedestrians and L = 20 for vehicles). The mean of the aspect ratios of the image instances (samples) with the specific quantized label is used as an aspect ratio for which a specific detector model is trained on that specific image instances.

\textit{Visual features}: We use visual features by clustering them as a means of capturing appearance variations of objects to detect them under challenging appearance changes. This approach is very generic as it does not depend on the availability of ground truth orientation though it gives slightly less accuracy when compared to 3D geometrical orientation.
Though color and gradient features (HOG, LUV color and normalized gradient magnitude) can also be used~\cite{OhnTri15}, in our case, high quality convolutional neural network (CNN) features from a R-CNN object detector~\cite{GirJef14} are used to learn appearance variations of objects. The R-CNN object detector model is then used to extract 4096-dimensional CNN features from cropped KITTI image samples which is then reduced dimensionally using PCA. This dimensionally reduced features are then clustered using k-means clustering giving cluster labels. The mean of the aspect ratios of the image instances assigned the same label is used to learn a specific detector on the image instances with that specific label.

\textit{Pedestrian Detection}: The ACF pedestrian detector~\cite{DolAppPerBel14} detects pedestrians at multiple scales using the Adaboost classifier. However, unlike the original ACF~\cite{DolAppPerBel14}, we consider the appearance variations of pedestrians due to varying orientation, truncation and occlusion levels. The KITTI benchmark~\cite{GeiLenStiUrt2013} consists of 7481 training frames from which around 3583 pedestrian instances are extracted in the moderate setting. From these 7481 training frames, 6501 frames are used as the training set and the rest (980 frames) as the testing or validation set. To capture the appearance variations, we trained 5 geometrical orientations-based detection models and 5 visual CNN features clustering-based detection models which perform better than using only one model as in~\cite{DolAppPerBel14}. This is necessary as using only one model, it is not possible to get the required detection performance for extracting detection probabilities on this data set. Overlapping detections are merged using a greedy NMS overlap threshold of 0.1. However, when evaluating the detector, an overlap threshold of 0.5 is used to identify true positives vs false positives.

\textit{Vehicle Detection}: Capturing appearance variations of vehicles due to changing observation angle, illumination variability, vehicle shape and type, truncation, out of camera view, different occlusion levels, etc is very important for developing a vehicle detector~\cite{OhnTri15}. Unlike the approach considered in~\cite{OhnTri15}, we consider only 3D orientation rather than other geometrical features such as truncation level, occlusion level and occlusion type features from the ground truth available in KITTI. Moreover, we used visual features which can also capture appearance variations due to varying orientation, truncation and occlusion. The KITTI benchmark~\cite{GeiLenStiUrt2013} consists of 7481 training frames from which around 16105 car instances are extracted in moderate setting. From these 7481 training frames, 6501 frames are used as the training set and the rest (980 frames) as the testing or validation set. Accordingly, we trained 20 geometrical orientations-based detection models and 20 visual CNN features clustering-based detection models which perform better than using only one model. We use a greedy NMS overlap threshold of 0.2 to merge overlapping detections, and an overlap threshold of 0.5 is used to identify true positives vs false positives when evaluating the detector.

\textit{Detection Parameters Extraction}: The detection parameters, detection probabilities ($p_{11,D}$, $p_{22,D}$) and confusion detection probabilities ($p_{12,D}$, $p_{21,D}$), are extracted as follows. The detection probability for pedestrians by a pedestrian detector, $p_{11,D}$, can be extracted from the ROC curve of the pedestrian detector when it is tested on pedestrian instances. Similarly, the detection probability for vehicles by a vehicle detector, $p_{22,D}$, is obtained from the ROC curve of the vehicle detector when it is tested on vehicle instances. The confusion detection probabilities, detection probability for pedestrians by a vehicle detector, $p_{21,D}$, and detection probability for vehicles by a pedestrian detector, $p_{12,D}$, can also be obtained when the vehicle detector is evaluated on pedestrian instances and when the pedestrian detector is evaluated on vehicle instances, respectively.

\begin{figure} [h] 
  \begin{center}
   \subfloat[\scriptsize{ROC for pedestrian detector on pedestrian instances}]
  {\label{fig:ROCgcnnPs16} \includegraphics[height=0.19\textwidth]{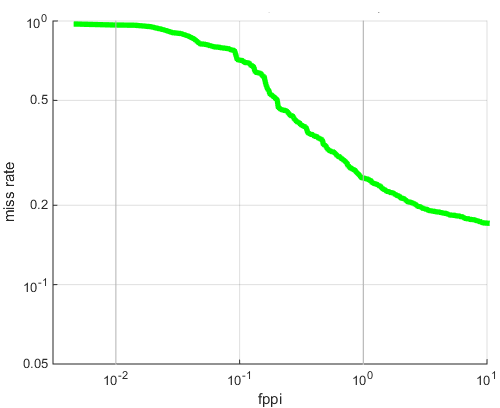}} 
   \subfloat[\scriptsize{ROC for vehicle detector on vehicle instances}]
  {\label{fig:ROCgcnnVs16} \includegraphics[height=0.19\textwidth]{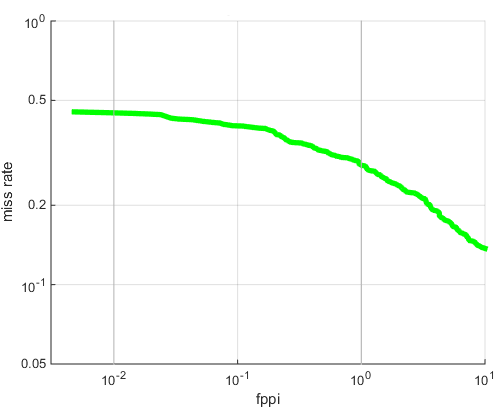}} 
  \end{center}
   \caption{\small{ROCs using 3D orientation and CNN visual features detector models tested on KITTI sequence 16.}}
  \label{fig:ROCgcnnPV}
\end{figure}

Accordingly, when we want to track on video sequences, we first run both pedestrian and vehicle detectors on that specific video sequence to obtain their ROC curves. We used a combination of CNN-based and 3D orientation-based detectors. Note that CNN features and 3D orientation are used for obtaining cluster labels; the core detector uses HOG, normalized gradient magnitude and LUV color features with Adaboost classifier. Thus, the ROC curve of the pedestrian detector when applied to pedestrian instances in the KITTI video tracking sequence 16 is shown in Fig.~\ref{fig:ROCgcnnPs16} from which $p_{11,D}$ of 0.83 is obtained at clutter rate (false positive per image - fppi) of 10. Similarly, $p_{22,D}$ of 0.86 is obtained at fppi of 10 from the ROC curve of the vehicle detector when it is applied to vehicle instances as shown in Fig.~\ref{fig:ROCgcnnVs16}. However, the values of $p_{12,D}$ and $p_{21,D}$ are very low, around 0.03 for $p_{21,D}$ and 0.01 for $p_{12,D}$, this happens because even if the vehicle detector detects pedestrian instances, for example, the intersection of the detected bounding boxes by vehicle detector on pedestrian instances and the ground truth of the pedestrian instances is very low as the two bounding boxes have very much different aspect ratios, therefore, it can be classified as false positive though it is detected. Hence, we try to fine-tune values of $p_{12,D}$ and $p_{21,D}$ to some higher values e.g. $p_{21,D} = 0.3$ and $p_{12,D} = 0.1$.

\noindent

\subsubsection{Data Association and Tracking Results}\label{Sec:trackingResults2}

Munkres's variant of the Hungarian assignment algorithm~\cite{FraJea71} is used to associate tracked target identities between two consecutive frames in the same fashion discussed in section~\ref{Sec:DataAssociation} with the exception that if a target disappears and then reappears, a new label is given without using any re-identification algorithm.

The state vector includes the centroid positions, velocities, and the width and height of the bounding boxes; the measurement is the noisy version of the target area, in the same fashion as in section~\ref{Sec:trackingResults}. A dynamic model, an observation model and a birth covariance follow Eqs.~(\ref{eqn:PHDstateTransitionMatrixVideo}), (\ref{eqn:eqn:PHDobservationMatrixVideo}) and (\ref{eqn:PHDbirthCovariance}) respectively with the exception of setting $i \in \{1, 2\}$. We set $\sigma_{v_1} = 5$ pixels$/s^2$  and $\sigma_{v_2} = 6$ pixels$/s^2$ for target type 1 (pedestrians) and target type 2 (vehicles), respectively. We also set survival probabilities $p_{1,S} = p_{2,S} = 0.99 $ for each target of both types, and the measurement standard deviations $\sigma_{r_{ii}}$ and $\sigma_{r_{ij}}$ ($i \in \{1,2\}$ and $j \in \{1,2\}$) are evaluated to 7 pixels.

\begin{figure*} [!htb] 
  \begin{center}
   \subfloat[\scriptsize{Detections, frame 13}]
  {\label{fig:DetectionFrame13} \includegraphics[width = 0.90\textwidth]{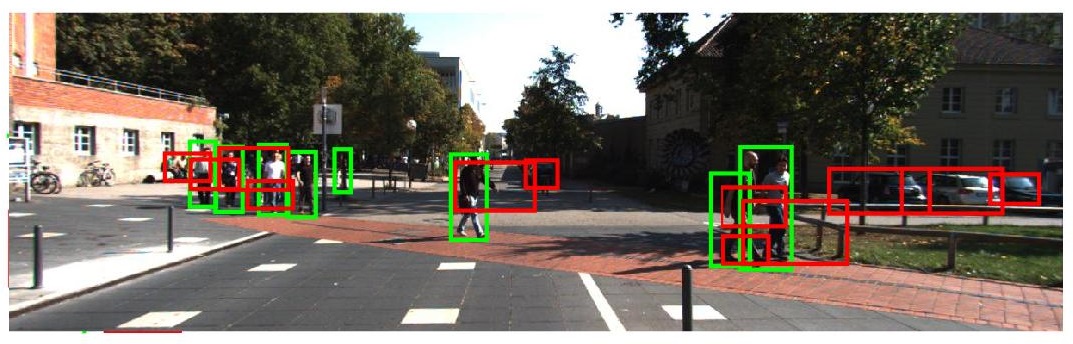}} \\ 
   \subfloat[\scriptsize{Two independent GM-PHD trackers, frame 13 }]
  {\label{fig:TwoPHDsFrame13} \includegraphics[width = 0.90\textwidth]{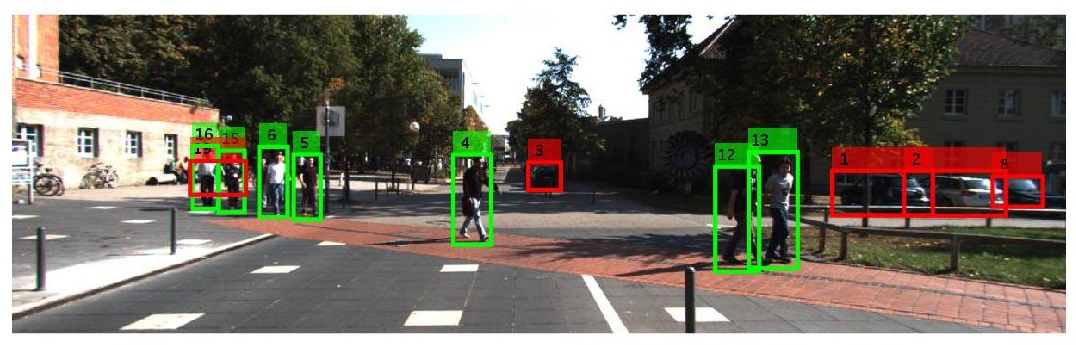}} \\
  \subfloat[\scriptsize{Dual GM-PHD Tracker, frame 13}]
  {\label{fig:DualPHDFrame13} \includegraphics[width = 0.90\textwidth]{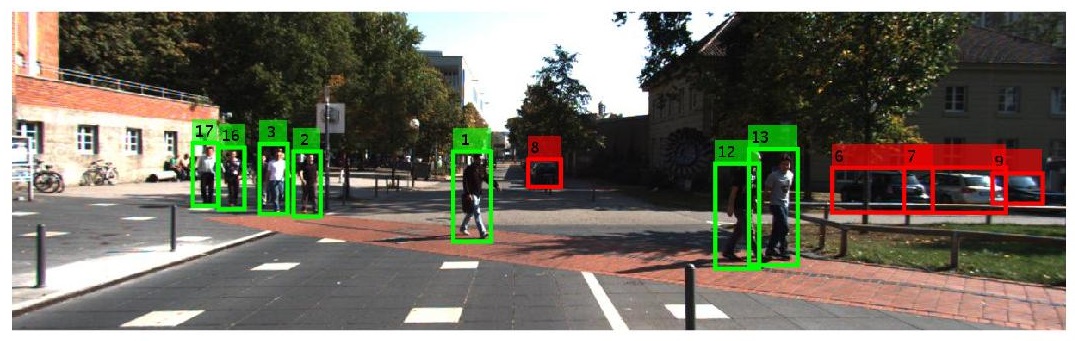}} \\
  \end{center}
   \caption{\small{Results of detections, two independent GM-PHD trackers and dual GM-PHD tracker, respectively, for frame 13.}}
  \label{fig:TrackingFrame13}
\end{figure*}

\begin{figure*} [!htb] 
  \begin{center}
   \subfloat[\scriptsize{Detections, frame 23}]
  {\label{fig:DetectionFrame23} \includegraphics[width = 0.90\textwidth]{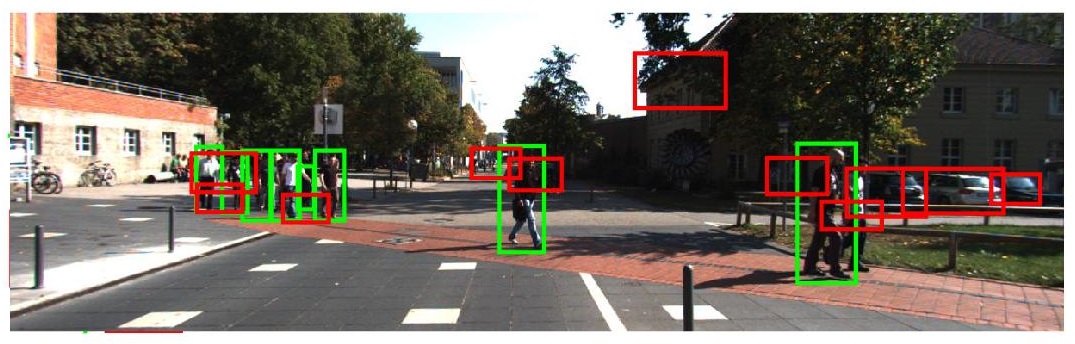}} \\ 
   \subfloat[\scriptsize{Two independent GM-PHD trackers, frame 23}]
  {\label{fig:TwoPHDsFrame23} \includegraphics[width = 0.90\textwidth]{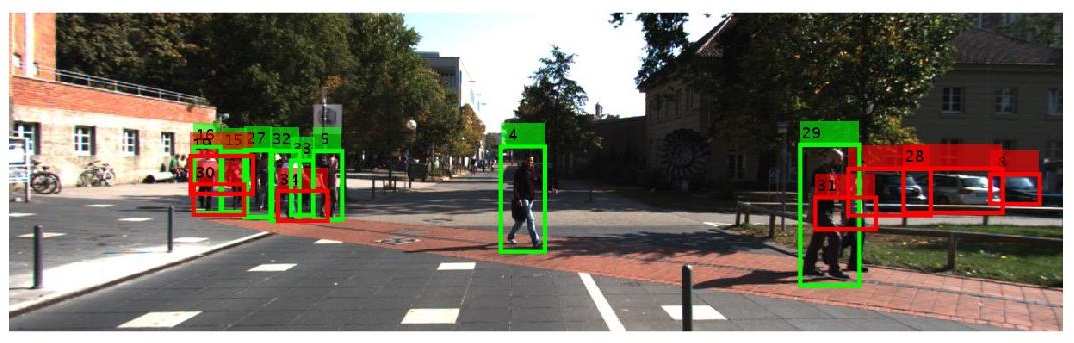}} \\
  \subfloat[\scriptsize{Dual GM-PHD Tracker, frame 23}]
  {\label{fig:DualPHDFrame23} \includegraphics[width = 0.90\textwidth]{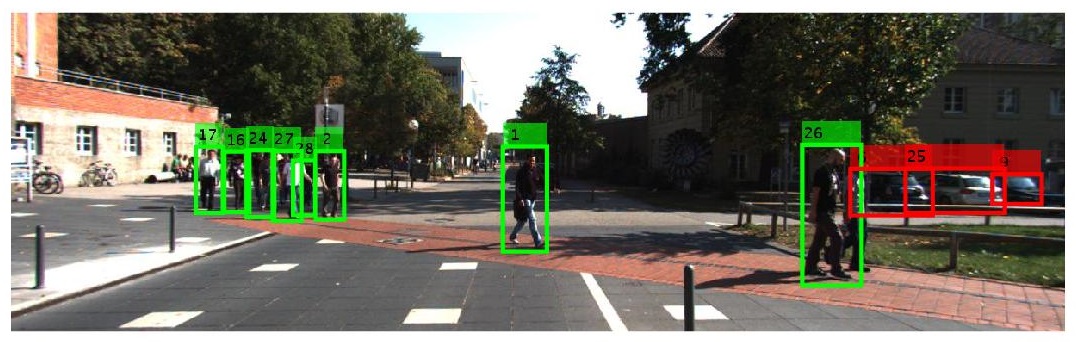}} \\
  \end{center}
   \caption{\small{Results of detections, two independent GM-PHD trackers and dual GM-PHD tracker, respectively, for frame 23.}}
  \label{fig:TrackingFrame23}
\end{figure*}

This proposed visual tracking approach is analyzed using the KITTI tracking video sequence 16. A multi-target measurement for pedestrians is obtained using a pedestrian detector and a multi-target measurement for vehicles is obtained using a vehicle detector. The sample frames of results of detections, two independent GM-PHD trackers and dual GM-PHD tracker are shown in Fig.~\ref{fig:TrackingFrame13} and Fig.~\ref{fig:TrackingFrame23}. For instance, for the sample frame 23 in Fig.~\ref{fig:TrackingFrame23}, many clutter responses from detections in Fig.~\ref{fig:DetectionFrame23} are removed by 2 independent GM-PHD trackers as shown in Fig.~\ref{fig:TwoPHDsFrame23}. However, the pedestrian targets with labels 10, 30, 34 and 31 are confused by the vehicle detector and then tracked by standard GM-PHD trackers as shown in Fig.~\ref{fig:TwoPHDsFrame23}. These are removed by our dual GM-PHD tracker as shown in Fig.~\ref{fig:DualPHDFrame23}. Hence, our approach eliminates the wrong tracking of vehicles or pedestrians which are confused at detection.

\begin{figure*} [!htb] 
  \begin{center}
   \subfloat[\scriptsize{Cardinality}]
  {\label{fig:VisCardinality2} \includegraphics[height=0.35\textwidth]{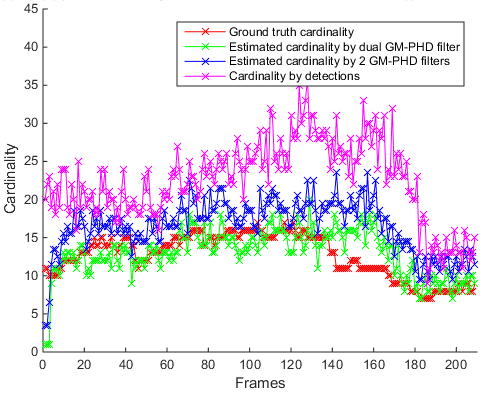}} 
   \subfloat[\scriptsize{OSPA metric error}]
  {\label{fig:VisOSPAerror2} \includegraphics[height=0.35\textwidth]{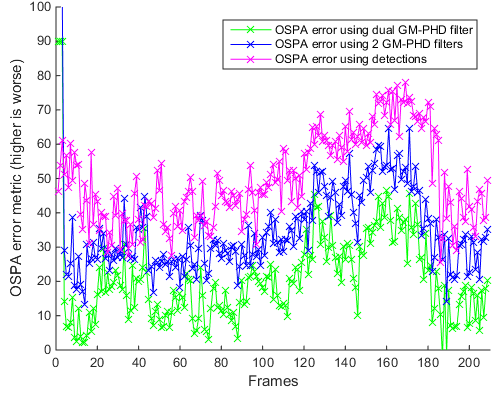}}
  \end{center}
   \caption{\small{Cardinality and OSPA error: Ground truth (red for cardinality only), dual GM-PHD filter (green), two independent GM-PHD filters (blue), detections (magenta).}}
  \label{fig:OSPA-CardinalityComparisonVis2}
\end{figure*}

The dual GM-PHD filter is evaluated quantitatively and compared with two independent GM-PHD filters and raw detection using the cardinality error, OSPA metric~\cite{SchVoVo08} (using order $p = 1$ and cutoff $c = 100$), time taken and discrimination rate in Table~\ref{tbl:OSPAerror2}. We also show the cardinality and OSPA error plots as shown in Fig.~\ref{fig:VisCardinality2} and Fig.~\ref{fig:VisOSPAerror2}, respectively, in red for ground truth (cardinality), green for dual GM-PHD filter, blue for two independent GM-PHD filters and magenta for detections. As shown in Table~\ref{tbl:OSPAerror2}, the overall average value of the OSPA error for the dual GM-PHD filter is 20.74 pixels compared to using two independent GM-PHD filters of 35.29 pixels and raw detection of 49.81 pixels.
Hence. our approach reduces the OSPA error by a large margin.

The time taken for the 209 video frames of the KITTI tracking sequence 16 is shown in Table~\ref{tbl:OSPAerror2} where the dual GM-PHD tracker takes 4.03 seconds per frame (both detection and tracking), two independent GM-PHD trackers take 1.61 seconds per frame and raw detection takes 1.25 seconds per frame on a i5 2.50 GHz core processor with 6 GB RAM laptop using MATLAB. Since we use many detection models for each actor (10 for pedestrians, 40 for vehicles), it takes more computational time than the scenario considered in section~\ref{Sec:threeTypes}. As shown in Table~\ref{tbl:OSPAerror2}, the dual GM-PHD tracker has only 0.32 (below 1 target) cardinality error and 1.81\% discrimination rate error when compared to 3.82 cardinality error and 100\% discrimination rate error using 2 independent GM-PHD trackers as well as 9.86 cardinality error and 100\% discrimination rate error using raw detection. As can be seen from Fig.~\ref{fig:DualPHDFrame13} and Fig.~\ref{fig:DualPHDFrame23}, labels of some of the actors (cars - labels 6 and 9; pedestrians - labels 1, 2, 16, 17, etc) are consistent from frame to frame. Since we are using two frames to associate the targets, a new label is given to a target which disappears and reappears as well as for a newly appearing target. For example, pedestrians labeled 12 and 13 in frame 13 are re-detected as one target in frame 23 and given a label 26. The car labeled 7 in frame 13 is mis-detected, and then is re-detected in frame 23 and is given a new label, 25. Still, the labeling approach we use has reasonable performance with a mean label switch error of only 1.07\%, and it is obviously not part of the dual GM-PHD filter.

\noindent

\noindent

\noindent


\begin{table*}[!htb]
\begin{center}
\begin{tabular}{|l|c|c|c|r|}
\hline
Method & cardinality error & OSPA error & time taken & discrimination rate \\
\hline\hline
Detections & 9.86 & 49.81 pixels & 1.25 sec/frame & 0\% \\
2 GM-PHDs  &  3.82 & 35.29 pixels & 1.61 sec/frame & 0\% \\
Dual GM-PHD &  0.32 & 20.74 pixels & 4.03 sec/frame & 98.19\% \\
\hline
\end{tabular}
\end{center}
\caption{\small{Frame-averaged cardinality and OSPA errors, time taken and discrimination rate at the extracted detection probabilities for dual GM-PHD filter, two independent GM-PHD filters and Detections.}}
\label{tbl:OSPAerror2}
\end{table*}

\subsection{Evaluation and Comparison using the MOT Benchmarking Tool} \label{Sec:MOTchallenge}

The N-type GM-PHD tracker is developed for handling confusions among detectors of different target types. However, all of the state-of-the-art multi-target tracking algorithms in the MOT challenge have been developed for tracking
multiple targets ($N=1)$ of a single type.
Thus, the N-type GM-PHD tracker cannot be directly evaluated using the MOT benchmarking tool to be compared to the state-of-the-art algorithms listed on the MOT challenge.
However, in response to comments, we use the building block of the N-type GM-PHD tracker (the N-type GM-PHD tracker for $N=1$, denoted by GM-PHD-N1T) to evaluate the performance of the MOT benchmarking tool and compare to the state-of-the-art algorithms in the MOT challenge.

Thus, we validate our tracker (for $N=1$ version) and compare it against state-of-the-art online and offline tracking methods (GM-PHD-HDA~\cite{SonJeo16}, DP-NMS~\cite{PirRamFow11}, SMOT~\cite{DicCamSzn13}, CEM~\cite{MilRotSch14} and JPDA-m~\cite{RezMilZha15}) on the MOT16 benchmark datasets~\cite{MOT16}. We use the \textit{public detections} provided by the MOT benchmark. We use the following evaluation measures: Multiple Object Tracking Accuracy (MOTA), Multiple Object Tracking Precision (MOTP)~\cite{BerSti08}, Mostly Tracked targets (MT), Mostly Lost targets (ML)~\cite{LiHua09}, Fragmented trajectories (Frag), False Positives (FP), False negatives (FN), Identity Switches (IDSw) and speed of the tracker (Hz). For a detailed description of each metric, please refer to~\cite{MOT16}.

Quantitative evaluation of our proposed method with other trackers is given in Table~\ref{tbl:MOT16}. The Table shows that our algorithm outperforms both online and offline trackers listed in the table in terms of MOTP, MOTA and FP; our algorithm provides the 2nd lowest FP of all listed algorithms on MOT 16 benchmark website. The number of MT percentage is overall higher than many of the online and offline trackers except one offline tracker (i.e. second to CEM). The number of ML and FN percentage are also lower than many of the online and offline trackers. The higher number of IDSw and Frag compared to the other online tracker and some of the offline trackers is due to the fact that our tracker, GM-PHD-N1T, relies only on the position and size of the bounding box of the detections; we are not using any appearance models to discriminate nearby targets in this tracker. Spawning targets are also not modelled in our tracker, therefore, identity switches are more likely to occur in such crowded scenes. For comparison, we also include appearance features, re-identification algorithm~\cite{EjaMic15}, to re-identify the targets in the last 5 frames for computational efficiency to re-assign their unique labels in case they disappear and then reappear. We designate this tracker as GM-PHD-N1F in the Table~\ref{tbl:MOT16}. Including appearance features (re-identification) improves the tracker's performance over using only position and size of the bounding box of the detections (GM-PHD-N1F vs GM-PHD-N1T) in terms of many metrics such as MOTA, IDSw and Frag. The GM-PHD-N1T tracker runs about 9.9 frames per second (fps) whereas GM-PHD-N1F runs about 8.9 fps. The computational costs arise from experiments on a i7 2.30 GHz core processor with 8 GB RAM using Matlab (not well optimized).

The most relevant comparison of our algorithm is to the GM-PHD-HDA~\cite{SonJeo16}. Both these trackers (GM-PHD-HDA and ours) use a GM-PHD filter but with different approaches to labelling of targets from frame to frame. While our trackers, both GM-PHD-N1T and GM-PHD-N1F, use the Hungarian algorithm for labelling of targets by postprocessing the output of the filter, the GM-PHD-HDA uses the approach in~\cite{PanClaVo09} by also including appearance features for labelling targets. Our trackers outperform the GM-PHD-HDA tracker in many of the evaluation metrics, but not IDSw and Frag. The GM-PHD-HDA performs better in terms of IDSw and Frag even if we include appearance features in addition to the position and size of the bounding box of the detection (GM-PHD-N1F).
However, our method outperforms it significantly in all other metrics.
Thus, when considered overall the ($N=1$) adaptation of the N-type GM-PHD tracker performs reasonably well
when evaluated and compared to the MOT challenge, even though the primary purpose of the algorithm is to consider different types.

\begin{table*}
\begin{center}
   \begin{tabular}{|l|c|c|c|c|c|c|c|c|c|r|}
   \hline
    Tracker & Tracking Mode & MOTA$\uparrow$ & MOTP$\uparrow$ & MT (\%)$\uparrow$ & ML (\%)$\downarrow$ & FP$\downarrow$ & FN$\downarrow$ & IDSw$\downarrow$ & Frag$\downarrow$ & Hz$\uparrow$ \\
    \hline
    CEM~\cite{MilRotSch14} & offline & \underline{33.2} & 75.8  & \textbf{7.8} &	\underline{54.4} &	6,837 &	\underline{114,322} & 642 &	\underline{731} & 0.3	\\
    DP-NMS~\cite{PirRamFow11} & offline & 26.2 &	\underline{76.3} & 4.1 & 67.5 &	\underline{3,689} &	130,557 & \textbf{365} & \textbf{638} & \textbf{212.6}  \\
    SMOT~\cite{DicCamSzn13} & offline & 29.7 & 75.2 & 5.3 & \textbf{47.7} &	17,426 & \textbf{107,552} & 3,108 & 4,483 & 0.2	 \\
    JPDF-m~\cite{RezMilZha15} & offline & 26.2 &	76.3 & 4.1 &	67.5 &	\underline{3,689} &	130,549 & \textbf{365} & \textbf{638} & \underline{22.2}	\\
    GM-PHD-HDA~\cite{SonJeo16} & \textbf{online} & 30.5 & 75.4 &	4.6 & 59.7 & 5,169 & 120,970 & \underline{539} & \underline{731} & 13.6 \\
    \textbf{GM-PHD-N1T (ours)} & \textbf{online} & \textbf{33.3} & \textbf{76.8} & \underline{5.5} & 56.0 & \textbf{1,750} & 116,452 & 3,499 & 3,594 & 9.9 \\
    \textbf{GM-PHD-N1F (ours)} & \textbf{online} & \textbf{33.8} & \textbf{76.8} & \underline{5.6} & 55.0 & \textbf{1,748} & 116,451 & 2,981 & 2,983 & 8.9 \\
    \hline
\end{tabular}
\end{center}
\caption{Tracking performance of representative trackers developed using both online and offline methods. All trackers are evaluated on the test dataset of the MOT16~\cite{MOT16} benchmark using public detections. The first and second highest values are highlighted by bold and underline, respectively.}
\label{tbl:MOT16}
\end{table*}

\section{Conclusions} \label{Sec:Conclusion}

We have developed an extension of the PHD filter in the RFS framework to account for $N\geq2$ different types of multiple targets with separate observations in the same scene, allowing for different probabilities of detection, scene clutter and possible confusions between targets of different types at the detection stage. This extends the standard GM-PHD filter~\cite{VoMa06} to a N-type GM-PHD filter.
This has been tested and evaluated using video sequences with the separate targets defined as different team players and the referee, and pedestrians and vehicles. We have also applied Munkres's variant of the Hungarian assignment algorithm as data association on the filtered results of the filter as a post-process. The key finding of this work is that by considering and modeling confusions between the different types of target at the detection stage we can improve the target discrimination rate, demonstrated by quantitative measurement of cardinality and the OSPA score.
Although the process has been applied here to $3$ and $2$ types of target, in principle the methodology can be applied to $N$ types of targets where $N$ is a variable, with the caveat that the number of possible confusions may rise as $N(N-1)$.

The N-type GM-PHD filter degrades to $N$ GM-PHD filters when we set the probabilities of confusion to 0.0 i.e. no target confusions. However, if each target is regarded as a type, the N-type GM-PHD filter is used as a labeler of each target i.e. it discriminates those targets from frame to frame whether or not confusions between targets exists rather than simply degrading to $N$ standard GM-PHD filters.
We also observe that other assumptions about clutter, target location and birth follow the same random models as the standard PHD filter.
Hence we assume that our background clutter, and the detection and confusion probabilities are uniform across the image field, which is not unreasonable in the football data, but is less likely to be true when identifying pedestrians in an urban environment, where particular street furniture may generate repeated false alarms. As the detections are represented as points (centroids of bounding detection boxes), the filtering process does not explicitly consider scale, and as the boxes and humans/vehicles within the boxes have finite extent, this makes occlusions possible such that targets may disappear for several frames. Notwithstanding these imperfections, this work has shown that the N-type GM-PHD filter has potential both to track targets in video data, and to better address multiple target confusions than the standard method.

\section*{Acknowledgment}

We would like to acknowledge the support of the Engineering and Physical Sciences Research Council (EPSRC), grant references EP/K009931 and a James Watt Scholarship. We would also like to thank Dr. Daniel Clark for sharing his expertise and understanding of RFS methodology.

\ifCLASSOPTIONcaptionsoff
  \newpage
\fi

\bibliographystyle{IEEEtran}
\bibliography{egbib}




\end{document}